\documentclass{article}

\usepackage{arxiv}

\usepackage[utf8]{inputenc}
\usepackage[T1]{fontenc}
\usepackage{hyperref}
\usepackage{url}
\usepackage{booktabs}
\usepackage{amsfonts, amsmath, amssymb}
\usepackage{graphicx}
\usepackage{subfigure}
\usepackage{enumitem}
\usepackage{natbib}
\usepackage{float}
\usepackage{array}

\title{Commute Networks as a Signature of Urban Socioeconomic Performance: Evaluating Mobility Structures with Deep Learning Models}

\author{
  Devashish Khulbe\thanks{Corresponding author: dk3596@nyu.edu} \\
  Department of Mathematics and Statistics\\
  Faculty of Science, Masaryk University \\
  \And
  Alexander Belyi \\
  Department of Mathematics and Statistics\\
  Faculty of Science, Masaryk University \\
  \And
  Stanislav Sobolevsky \\
  Center for Urban Science + Progress\\
  New York University \\
  Department of Mathematics and Statistics\\
  Faculty of Science, Masaryk University \\
}

\begin{document}
\maketitle

\begin{abstract}
Urban socioeconomic modeling has predominantly concentrated on extensive location and neighborhood-based features, relying on the localized population footprint. However, networks in urban systems are common, and many urban modeling methods don't account for network-based effects. In this study, we propose using commute information records from the census as a reliable and comprehensive source to construct mobility networks across cities. Leveraging deep learning architectures, we employ these commute networks across U.S. metro areas for socioeconomic modeling. We show that mobility network structures provide significant predictive performance without considering any node features. Consequently, we use mobility networks to present a supervised learning framework to model a city's socioeconomic indicator directly, combining Graph Neural Network and Vanilla Neural Network models to learn all parameters in a single learning pipeline. Our experiments in 12 major U.S. cities show the proposed model outperforms previous conventional machine learning models. This work provides urban researchers methods to incorporate network effects in urban modeling and informs stakeholders of wider network-based effects in urban policymaking and planning.
\end{abstract}





\section{Introduction}

Interactions among urban neighborhoods are increasingly common in the modern world. The interactions can take the form of physical movement of people and things or manifest in intangible quantities like social networks. A location's characteristics may be determined by its interactions with other places within the city.
Evaluating a region's place in a large urban network could determine many important indicators. This could reveal the social well-being of the residents, as well as the general economic status of a place in a city. Networks based on urban systems' geographical, cultural, and social perspectives have been well established and studied~\cite{NetworkInCities1, NetworkInCities2}. In particular, urban street-based networks are widely evaluated in the literature \cite{NetworkInCities3, NetworkInCities4}, and their impact on socioeconomics has also been linked~\cite{NetworkInCities5}. However, studying networks in the context of socioeconomic modeling is not very common in the literature. Urban scientists have focused more on aggregated physical quantities \cite{SocioeconomicModels1} and contextual local features \cite{SocioeconomicModels2} in socioeconomic modeling. Extensive urban variables like 311 service requests have proved to be crucial signatures of an urban neighborhood~\cite{SocioeconomicModels3}. However, local regional variables are only able to capture the information within a location. Traditional feature-based models often overlook structural and relational dependencies between urban regions, despite the fact that such mobility-driven network effects—reflecting how regions are interconnected through commuting flows—can implicitly influence where people choose to live, work, or relocate within a city. Hence, it has become increasingly crucial to have urban network representations as a key factor in different aspects of urban modeling. Such representations could capture intricate relations and topology of a region within a vast graph. Moreover, evaluating predictive models based on these representations with/without contextual regional variables would help determine the explanatory power of just the network in urban modeling. This could be quite helpful to urban scientists who rely on static regional features that are not updated frequently, while many urban networks can be constructed in almost real-time. Leveraging urban networks in socioeconomic modeling could thus be a crucial way to model city dynamics efficiently. 

Representation methods to generate low-dimensional embedding in a graph have existed in the literature. Methods such as node2vec \cite{node2vec} and LINE \cite{LINE} have been used to capture the structural properties of networks. These methods have limitations in capturing the global topology of a graph as they rely on local structures. Recently, Graph Neural Networks (GNN) \cite{GNN} have shown promising results in graph representation experiments for many downstream tasks like node classification \cite{GNN} and community detection \cite{GNN_commDetect}. GNNs essentially work on the principle of propagating information through convolution operations within a network, with the goal of preserving important structural properties of the network, such as local connectivity and symmetry. The convolution operation is implemented with Graph Convolutional Networks (GCN), while the attention mechanism has also been incorporated in the form of Graph Attention Networks (GAT) \cite{GATVelickovic}. We will hereby include GCN and GAT under the umbrella term of GNN.

Recent work in network representation modeling has also extended to urban networks, although in a limited way. Urban networks are typically constructed by defining nodes as geographic regions—such as grid cells, administrative zones, or neighborhoods—and edges as interactions or relationships between these regions. These interactions may include commuting flows, taxi trip counts, social connections, or even co-occurrence of events, depending on the application context. This spatially grounded formulation enables the encoding of urban dynamics and functional connectivity patterns into graph structures that can be exploited by GNNs. Machine learning techniques like Variational Autoencoders have been used to represent urban street networks and compare their metrics \cite{NetRepr_streetNetwork}. For street networks, researchers have also proposed using different path (edge) representations for predictive modeling \cite{NetRepr_streetNetwork2}. In the context of node representation GATs have shown promising results for learning embedding for regions in an urban network \cite{NetRepr_GAT1, NetRepr_GAT2}. GATs have been used in an unsupervised learning fashion to learn the vector embedding with initial Point-of-interest (POI) features as input \cite{NetRepr_GAT1}. POI features have also been considered with GAT in the case of heterogeneous graphs, where researchers have used multiple node categories to construct the urban network \cite{NetRepr_GAT2}. More recently, GAT \cite{GATVelickovic} has been used to capture population-facilities interactions and dependencies in a bipartite urban graph \cite{GAT_population_facilities}. GNN-based node embedding vectors have shown promising results in supervised classification tasks, such as classifying building patterns based on building footprints \cite{GNN_classification1} and classifying urban scenes incorporating both visual and semantic features \cite{GNN_classification2}. GNNs are being increasingly applied in policy-relevant contexts. Recent methods have provided effective frameworks for traffic simulation \cite{natterer2024graphneuralnetworkapproach}, road-safety sensing \cite{zhang2020multi}, and land-use inference \cite{zhai2024heterogeneousgraphneuralnetworks}.

In the domain of socioeconomic modeling, recent work has demonstrated the potential of graph convolutional network (GCN)-based embeddings—trained to reconstruct network edges—for predicting variables such as median income and housing characteristics across city districts \cite{GNN_mobilityNet}. These approaches have achieved promising results by leveraging the structural information of urban networks; however, they typically rely on heterogeneous graphs that incorporate multiple categories of points of interest (POIs) and their links to spatial regions. Such models thus depend heavily on rich auxiliary datasets and fine-grained contextual information, including POI annotations and regional attributes. Furthermore, many existing studies integrate extensive node-level features during training, often overlooking the question of how much predictive power lies in the network structure itself. This raises a critical research question: To what extent can urban networks alone, without additional contextual data, support reliable and interpretable socioeconomic modeling? By isolating the structural signal in mobility networks, we aim to assess whether they suffice for modeling key socioeconomic indicators, or if context-dependent features are indispensable for accurate predictions. 

In applications to downstream modeling, most existing approaches with GNNs follow a two-stage pipeline: first, a model is trained to generate node embeddings from the network—either in a supervised or unsupervised fashion—and then a second supervised model is trained to predict target variables using these learned embeddings as input features~\cite{NetRepr_GAT1, NetRepr_GAT2, GNN_mobilityNet}. While effective in some cases, this decoupled learning framework introduces a disconnect between representation learning and the end prediction task. It also requires careful tuning of two separate models with distinct objectives—embedding quality and predictive accuracy—which may not align optimally. This presents a notable gap: the absence of end-to-end models that directly learn to predict socioeconomic outcomes from the network structure. A more integrated approach, where the model is trained directly on the target variable using graph-based signals, could offer improved performance and interpretability, particularly in urban contexts where network structure may already encode meaningful socioeconomic patterns. Exploring whether such direct modeling approaches outperform two-stage pipelines is a key direction for advancing socioeconomic inference from urban mobility networks.

Our main contributions can be summarized as follows:

\begin{itemize}
\item We demonstrate that mobility (commute) networks alone, without reliance on auxiliary contextual data, can provide sufficient structural signal for effective socioeconomic modeling.
\item We propose a unified GNN+VNN architecture that jointly learns network embeddings and performs downstream prediction of urban location characteristics in a single end-to-end training pipeline.
\end{itemize}

\section{Materials and Methods}

\subsection{Data overview}
\begin{figure}[h]
    \centering
    \subfigure[LEHD commute flow network]{
        \includegraphics[width=0.43\textwidth]{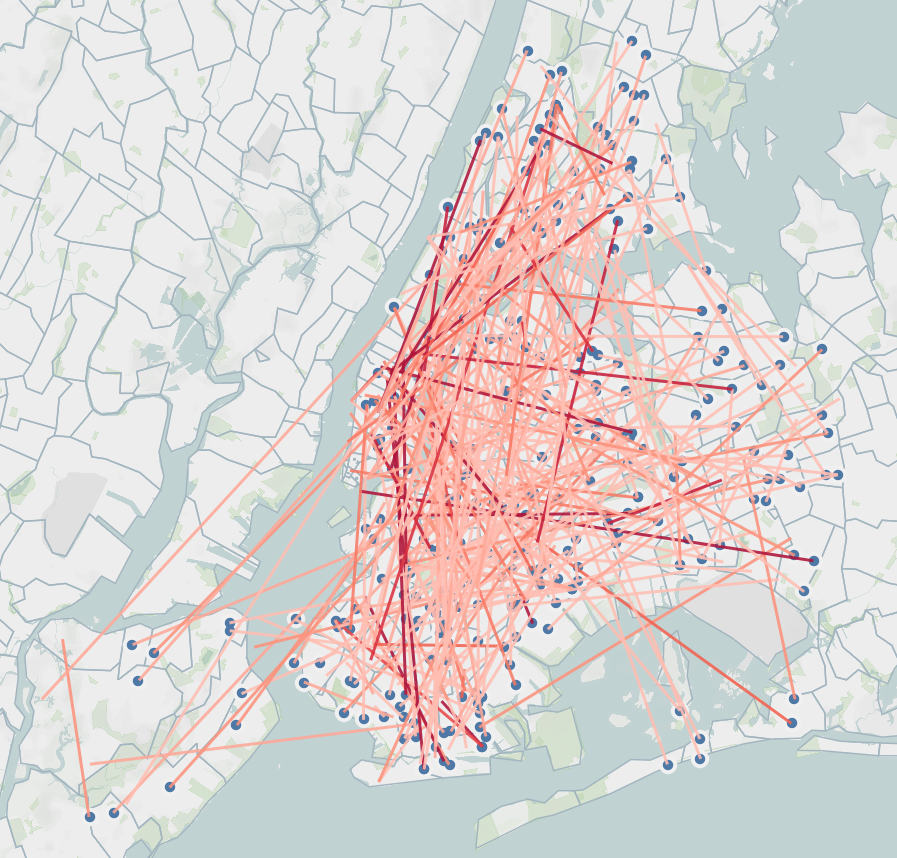}
        }
    \subfigure[Network structure and median income data for NYC]{
        \includegraphics[width=0.45\textwidth]{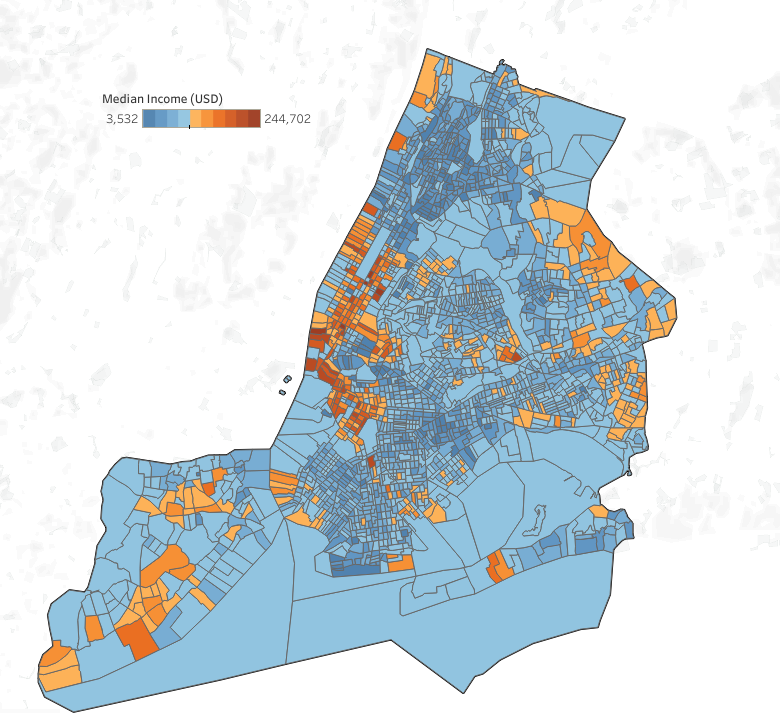}
        }
    \caption{Median income distribution across NYC census tracts}
    \label{dataFig}
\end{figure}

We define the urban mobility network by the commute flow among a city's neighborhoods. Specifically, we consider the network with nodes to be the geographical units in a city with edges being weighted by the number of people who commute for work from one unit to another per day. The commute flow data is retrieved from the Longitudinal Employer-Household Dynamics~(LEHD), a U.S. Census Bureau program \cite{LEHDData}. LEHD creates a detailed picture of labor market dynamics and integrates federal, state, and Census Bureau data on employers and employees. The commute flows are specifically collected by LEHD Origin-Destination Employment Statistics (LODES), which is updated annually. We use the commute flow values to populate our networks as they reflect a comprehensive picture of mobility in a city. While the LEHD data may not be as dynamically updated as other information about a city, it nevertheless represents an inclusive mobility network when contrasted with urban networks built on data from platforms like social media, which may exclude significant portions of the population. Table ~\ref{tab:city_stats} shows the key network statistics for the three cities. For the socioeconomic variable, we consider median income as the quantity for modeling. While income may not be a complete measure of the socioeconomic status of a neighborhood, it captures essential socioeconomic features. Moreover, other variables like unemployment status, housing profile, etc. are not easily available in many urban areas. We retrieve the income data for the cities from American Community Survey (ACS) data by U.S. Census \cite{ACSData}, which is updated annually. All the data is aggregated on the census tract level. Fig.~\ref{dataFig} shows the network structure and income distribution for NYC across census tracts. 

\begin{table}[!h]
  \caption{Commute network statistics for 12 cities}
  \label{tab:city_stats}
  \begin{tabular}{lccc} 
    \textbf{City} & \textbf{Nodes} & \textbf{Non-zero Edges} & \textbf{Avg. edge weight} \\
    New York & 2157 & 976832 & 0.69 \\
    Chicago & 1318 & 439553 & 1.06 \\
    Boston & 520 & 127357 & 3.5 \\
    Austin & 218 & 34777 & 8.63 \\
    Dallas & 529 & 129352 & 2.83 \\
    Los Angeles & 2341 & 1171362 & 0.65 \\
    San Antonio & 366 & 83192 & 4.77 \\
    San Diego & 627 & 180781 & 2.97 \\
    San Jose & 372 & 81938 & 4.73 \\
    Philadelphia & 384 & 68119 & 2.57 \\
    Phoenix & 916 & 349894 & 2.10 \\
    Houston & 786 & 290496 & 2.50 \\
  \end{tabular}
\end{table}

For comparison, we also consider the 311 complaint data set\cite{311Data} (based on availability) as features for an area. The data consists of all the non-emergency complaints across a city and comprises of a large number of complaint categories, thus making it a comprehensive representation of a given area's socioeconomic status. The results from 311 data-based modeling serve as a robust and proven benchmark \cite{SocioeconomicModels3} for comparison to our proposed methods.
More details on the 311 data are provided in Appendix:~\nameref{appendix:Data}.

\subsection{Methods}
We first investigate mobility network embeddings as predictors of a demographic target variable. The goal is to consider only the network structure within a city and evaluate its utility in modeling a socioeconomic indicator across the city's census tracts. In the following sections, we propose learning node representations using a VNN architecture to reconstruct the network edges. The VNN model follows a two-step learning process based on two MLP neural network models. In our next approach, we stack GNN and MLP layers to build a model that directly learns the socioeconomic target variable. This unified pipeline (hereafter referred to as the GNN+VNN framework) combines graph convolution operations with fully connected MLP layers to jointly learn all parameters while optimizing node embeddings for socioeconomic modeling.

\subsubsection{Network embedding as model inputs}\label{sec: intiEmbedding}
Graph embeddings have been extensively experimented with and used in graph-based models in the literature. In recent work in graph Transformer models (GT), network embeddings have been used as positional encodings. Some of the techniques for generating such embedding include SVD \cite{SVDencoding}, Laplacian Eigenvectors \cite{GraphTransformers, LaplacianEncoding1, LaplacianEncoding2}, and shortest path distances \cite{ShortestPathEncoding}. Recently, random walk encodings have been successfully used as structural encoding in GNN-based models \cite{RandWalkEncoding1, RandWalkEncoding2, encoding3}. In mobility networks, network embeddings have been proven to be useful for downstream tasks using heterogeneous networks \cite{UrbanEmbedding1, UrbanEmbedding2}. 

\begin{figure}[h]
    \centering
    \subfigure{
        \includegraphics[width=0.47\textwidth]{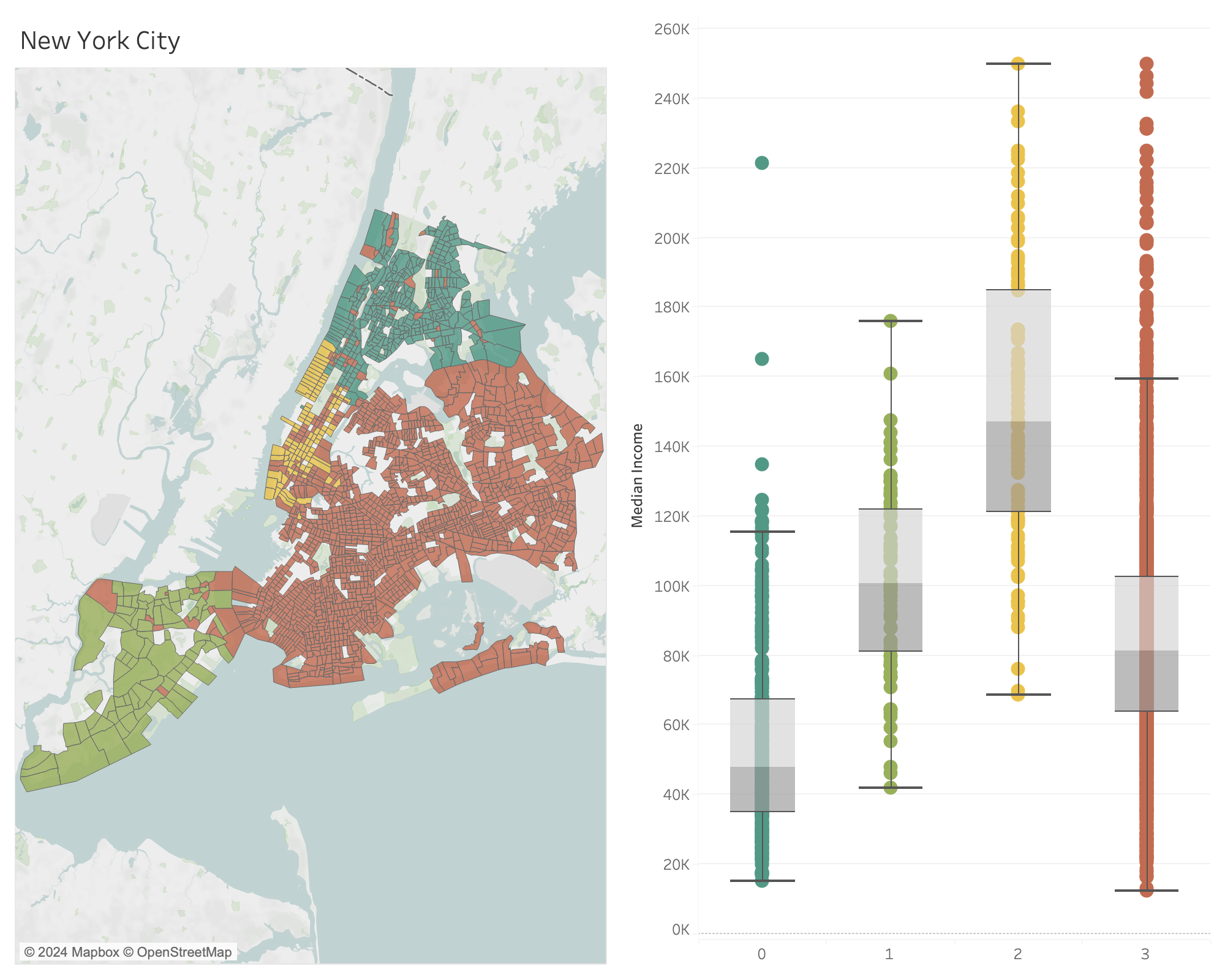}
        }
    \subfigure{
        \includegraphics[width=0.47\textwidth]{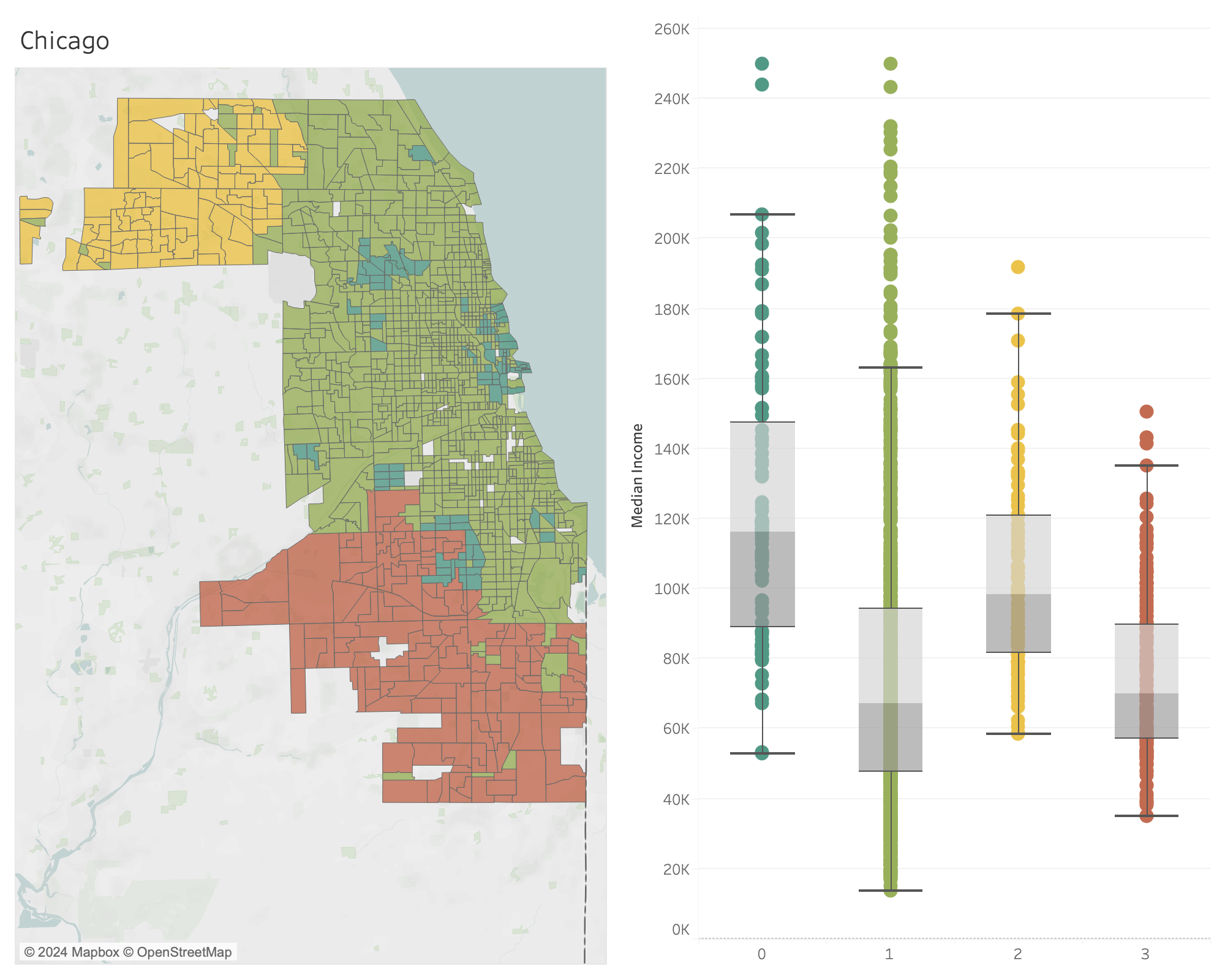}
        }
    \subfigure{
        \includegraphics[width=0.47\textwidth]{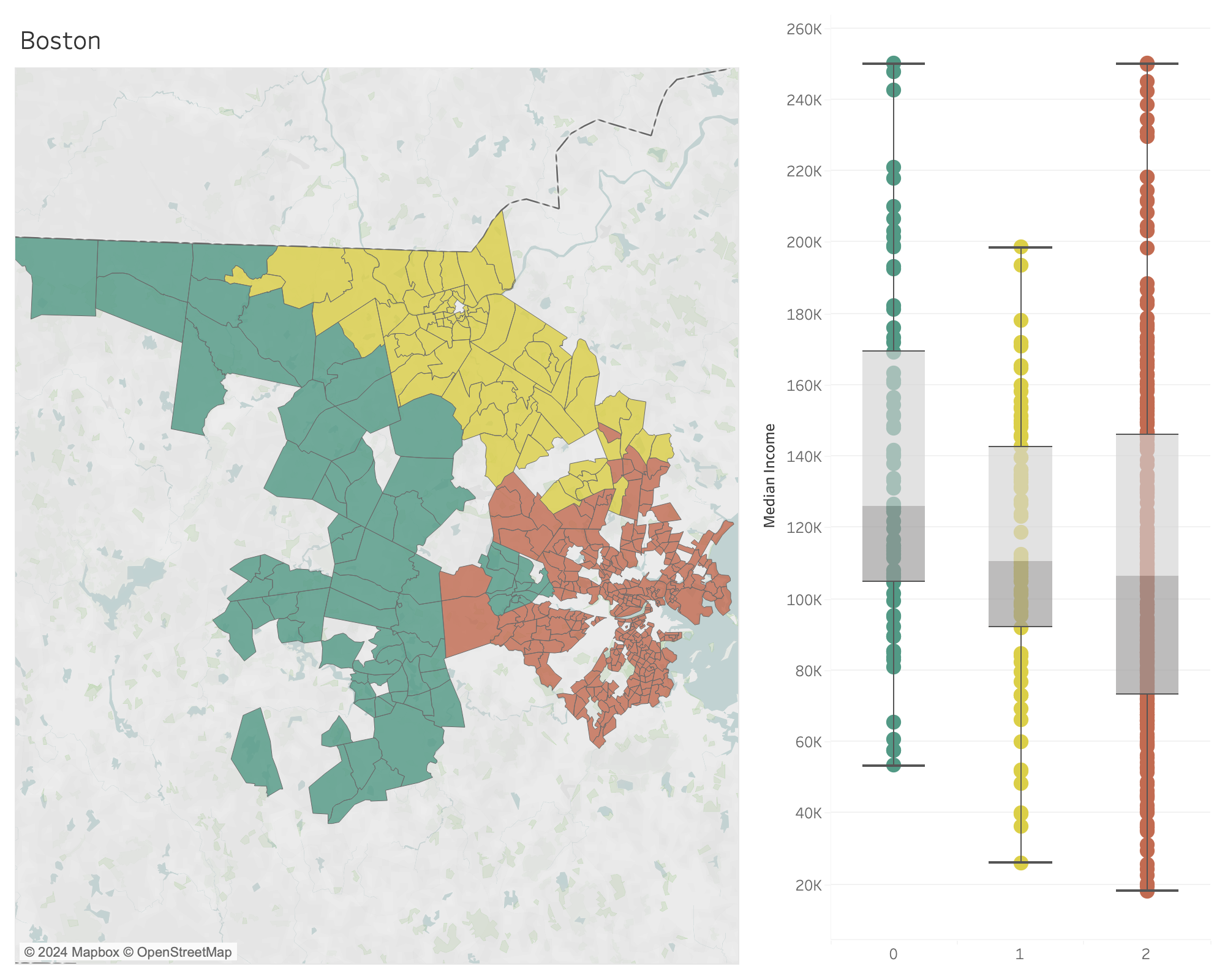}
        }
        \subfigure{
        \includegraphics[width=0.47\textwidth]{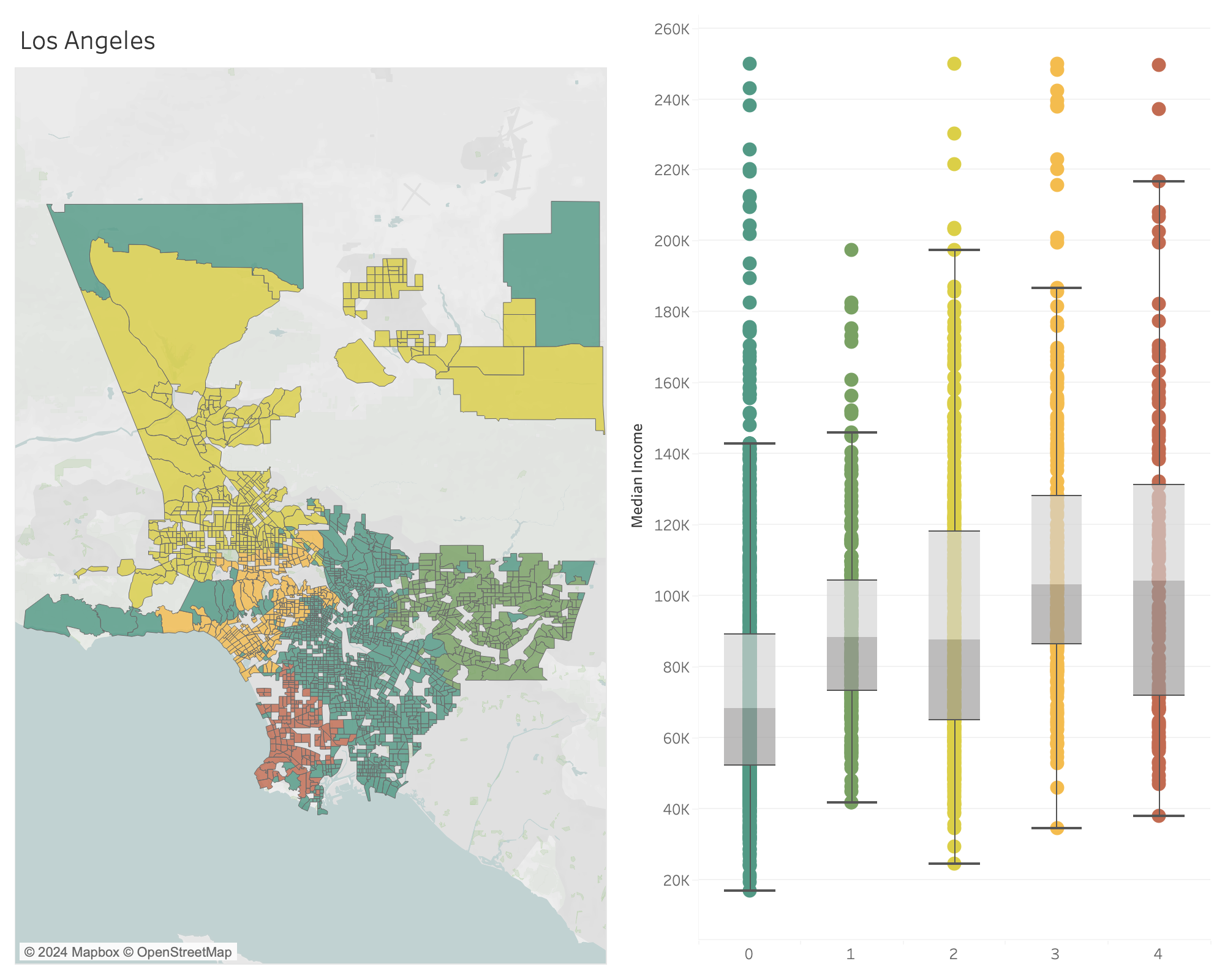}
        }
    \caption{\textbf{Clustering of census tracts in cities based on SVD embedding of mobility networks} -- spatial patterns reveal distinction of high-income from low-income neighborhoods}
    \label{fig:SVDclusters}
\end{figure}

Notably, in our experiments with clustering node embedding for mobility networks, we notice an interesting capability to distinguish regions based on their socioeconomic profile. In all cities considered in this work, we particularly observe that embeddings can differentiate between high-income and low-income districts in cities. Fig.~\ref{fig:SVDclusters} shows the results with K-means clustering model \cite{kmeans} on the SVD embedding. The median income of clusters can indicate the distinction of regions based on their socioeconomic profiles. Particularly, the clear distinction among areas such as lower Manhattan, the Bronx, inner Brooklyn/Queens in NYC, and Chicago's south side from central and north neighborhoods of the city is interesting. Clustering results from more embedding methods are discussed in Appendix:~\nameref{appendix:EmbeddingViz}

A recent study has found the potential to delineate regions in cities using cell phone mobility data \cite{cellPhoneDelineation}. The ability of mobility embedding to discern regions based on their socioeconomic profiles is particularly interesting in our experiments, as recent studies have also found that mobility is vastly impacted by commuters' income status \cite{IncomeCommute1, IncomeCommute2}. Thus, embedding as a low-dimensional representation of the larger mobility network can be particularly useful for modeling other socioeconomic indicators of regions, including income. We thus propose to use these embeddings as inputs in our models. Specifically, we experiment with spatial embedding (location of regions), SVD and Laplacian eigenvectors as initial inputs.

\subsubsection{Evaluating mobility networks~-- VNN based embedding}

\begin{figure}[!h]
\centering
\includegraphics[scale=0.3]{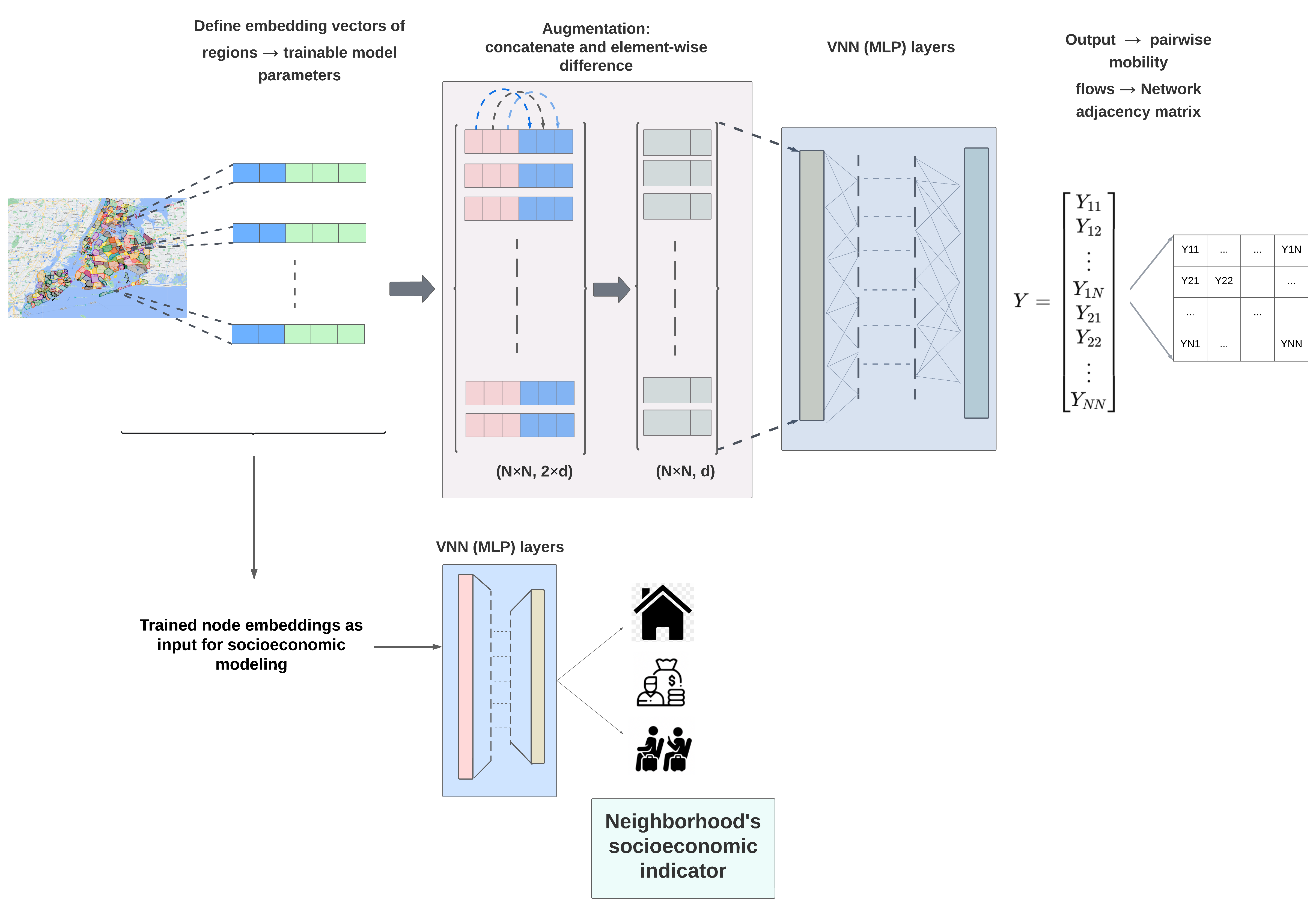}
\caption{
\textbf{VNN-based embedding model} -- Socioeconomic modeling is done in a two-step process: 1. Trainable regions' embeddings are processed and fed to an MLP to reconstruct the original O-D mobility flow matrix, 2. The learned embeddings are then fed as input to an MLP for modeling median income}
\label{MLPemb}
\end{figure}

We aim to learn vector representations of nodes (regions) in a city's mobility network, which can then be used as features for socioeconomic modeling. Urban network representations have been done using various methods, including deep learning-based architectures trained in supervised/unsupervised manner \cite{NetRepr_streetNetwork2, NetRepr_GAT1, GNN_classification1}. We begin by training a non-graph-based neural network as a baseline for comparison with the proposed graph-based models. Specifically, we use a VNN to reconstruct the edges of the mobility network, thereby learning structural representations of nodes. This training is conducted in a self-supervised manner, where the objective is to reconstruct the original adjacency structure of the network. The resulting internal states of the model serve as learned embeddings of the mobility graph, which can then be used for downstream task. Fig.~\ref{MLPemb} shows the model with its inputs and outputs involved. 

Let $G(V, E)$ be the mobility network with nodes $V$ representing census tracts and edges $E$ weighted by the volume of commute flow among the census tracts in a given city.
We consider an initial $d$ dimensional vector $e=[e_1, e_2,..., e_d]$ populated by the node embedding and random values, denoting the embedding for a given geographical entity in a city.
Then $E \in {N \times d}$ initializes the embedding matrix where each row represents the $d$ dimensional embedding vector for a specific entity (e.g., an area in the city). We then augment this matrix to transform it into a pairwise interaction matrix among the embedding pairs. The augmentation process involves creating a new matrix $E_{augmented} = concatenate([E_i, E_j]), E_i, E_j \in E$ by concatenating each row with every other row and itself from the original matrix. The concatenation is performed for all $i, j \in V$ (including $i=j$), accounting for self-loops in the network. We then compute the element-wise squared difference between the embedding vectors of paired entities to establish the interaction or dissimilarity measure between the two entities, which can be given as $(e'[:d]-e'[d:])^2, e' \in E_{augmented}$. The resulting matrix $E_{augmented}$ has dimensions 
$(N^2, d)$, capturing pairwise combinations of the original embedding vectors.

With the augmented matrix as the input, we consider a 3-layer VNN model with $(4\times d,3\times d,d)$ hidden layers ($d$ being the embedding dimensionality) and Rectified Linear Unit~(ReLU) activation. Specifically, the output of VNN model ($f$) is 

\begin{equation}
    Y_{ij} = f(E_{augmented}[ij])
\end{equation}
where the output $Y$ is a $N^2\times 1$ vector with a specific element $Y_{ij}$ representing the mobility between the $i_{th}$ and $j_{th}$ geographical entities. Importantly, we specify the individual nodes' vector (embedding) $E$ as a trainable parameter of the model. Therefore, the input vector can adapt during training to better represent the characteristics of the input data. Consequently, the model's output $Y$ can be reconstructed as the adjacency matrix $A$, which completely represents the connectivity between nodes in the network. The model is thus trained with the objective of reconstructing the adjacency matrix of the mobility network, with mean squared error~(MSE) as the objective to minimize:

\begin{equation}
\label{eq:1}
 \frac{1}{N^2}\sum_{i}^{N}\sum_{j}^{N}\left(A_{ij} - \hat{A_{ij}}\right)^2 
\end{equation}

Here, $A_{ij}$ represents the element at row $i$ and column $j$ in the true adjacency matrix A, while $\hat{A_{ij}}$ represents the corresponding element in the predicted adjacency matrix. 

The trained matrix $E$ can thus be regarded as an embedding of the network nodes. Each $d$-dimensional vector within this embedding corresponds to a specific region (node) within the mobility network, capturing its unique characteristics and interactions.

Modeling the socioeconomic variable with learned embedding as inputs with a supervised MLP model as a second step. The configuration of the model is discussed in \nameref{sec: Results} section, while performance with different embedding dimensions is discussed in Appendix:~\nameref{appendix:EmbedEx}

\subsubsection{Single pipeline modeling: GNN+VNN framework}

\begin{figure}[!h]
\centering
\includegraphics[scale=0.3]{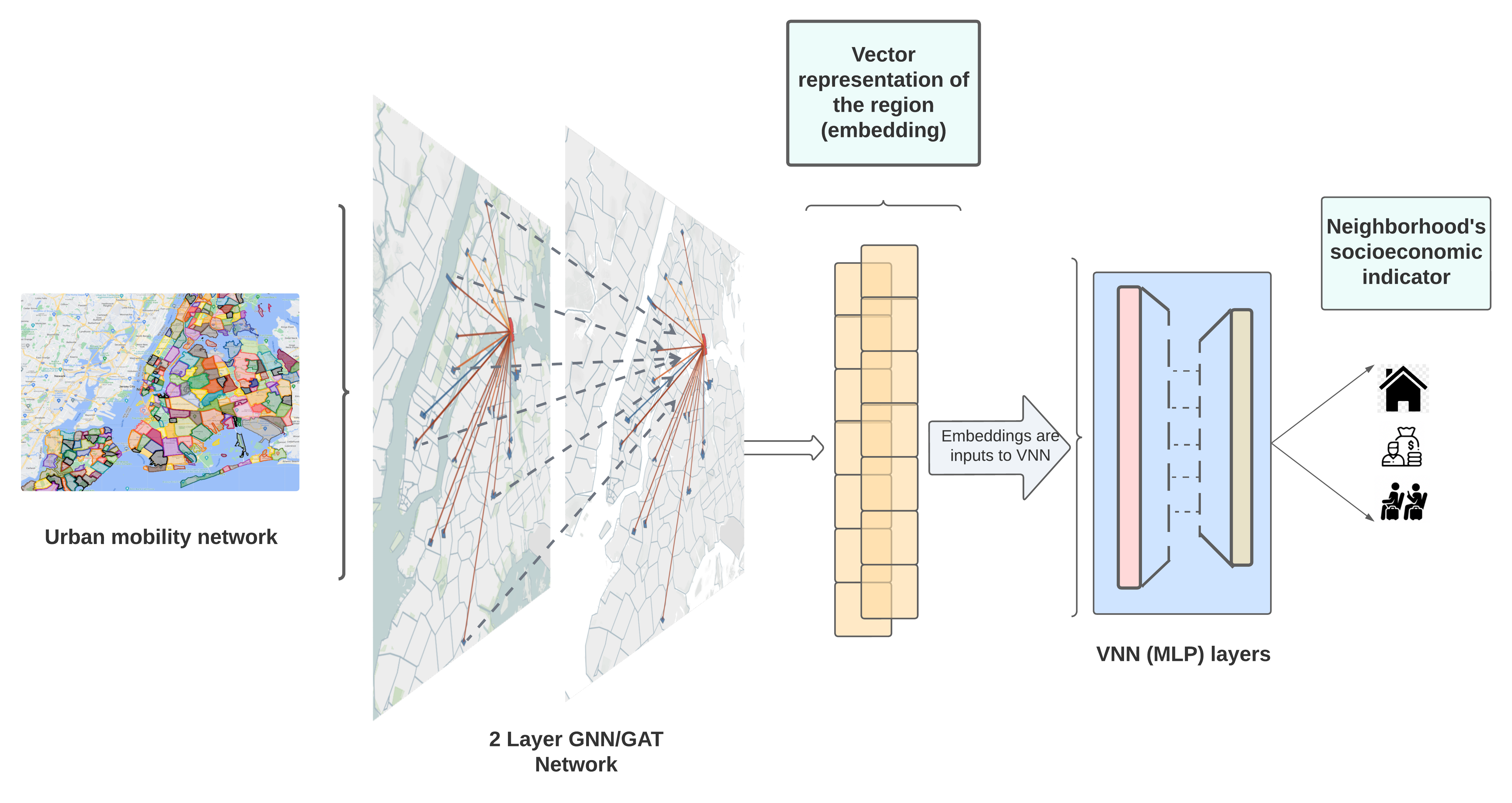}
\caption{
\textbf{Graph-based models} -- GCN/GAT layers are stacked with the MLP layers to directly model the socioeconomic feature}
\label{GNNemb}
\end{figure}

Using pre-trained network representations as features for various downstream tasks has been prevalent in the literature \cite{NetRepr_GAT1, NetRepr_GAT2, GNN_mobilityNet}. However, these methods require learning embeddings with a defined objective (e.g., edge reconstruction) which may have little significance to the overall objective in urban socioeconomic modeling. Hence, we propose a model architecture to learn the target variable in a single learning pipeline, without having to learn network embedding with a separate model. Fig.~\ref{GNNemb} depicts the complete model architecture with all layers involved.

We consider a 2 layer GNN model, considering the network adjacency matrix. A two-layer GNN architecture has been widely recognized as effective for various downstream tasks in graph-based learning \cite{GNN, GNN_2layer}. We leverage this proven configuration to capture the underlying structure and connectivity of nodes within the network. A typical GNN layer involves graph convolutions taking into account the topology and connectivity of nodes in the network. These layers operate by aggregating information from neighboring nodes and updating node representations accordingly. The graph convolution operation for a single layer can be defined as follows: 

\begin{equation}
    H^l = \sigma \left(W^l \hat{A} H^{l-1} + B^{l}\right)
\end{equation}

where $H^l \in R^{V \times d}$ is the output of the layer, $W^l$ is the weight matrix, $\hat{A} = D^{-1/2}AD^{-1/2}$ is the normalized adjacency matrix with self-loops ($D$ being the diagonal degree matrix), $B^{l}$ is bias vector at layer $l$ and $\sigma$ is an activation function. We consider ReLU as the activation in our configuration. Introducing normalization to account for the scale of sub-regions in the network, the convolution operation at the node level is given by:

\begin{equation}
    h_i^{l} = \sigma \left(\sum_{j\in N(i)}\frac{1}{|N(i)|}W^Th_j^{l-1}\right)
\end{equation}

where $N(i)$ is the connected neighborhood of node $i$, and $h_{i}^l$ being its feature vector in network embedding $H^l$.

While GCNs apply graph convolutions, the attention mechanism has been wildly successful in recent applications. The GAT model assigns different attention weights to neighboring nodes, thereby aggregating information in a more nuanced manner and updating node representations accordingly. The graph attention operation for a single head can be defined as follows:

\begin{equation}
h_i^{l} = \sigma \left(\sum_{j \in N(i)} \alpha_{ij} W^l h_j^{l-1} \right)
\end{equation}

where $h_i^{l} \in \mathbb{R}^d$ is the output feature vector of node $i$ at layer $l$, $W^l$ is the weight matrix, and $\alpha_{ij}$ are the attention coefficients, which are computed as:

\begin{equation}
\alpha_{ij} = \frac{\exp\left(\mathrm{LeakyReLU}\left(a^T [W^l h_i^{l-1} || W^l h_j^{l-1}] \right)\right)}{\sum_{k \in N(i)} \exp\left(\mathrm{LeakyReLU}\left(a^T [W^l h_i^{l-1} || W^l h_k^{l-1}] \right)\right)}
\end{equation}

where $a$ is a learnable weight vector, and $||$ denotes concatenation. The LeakyReLU activation function introduces non-linearity into the attention mechanism. To enhance the model's capability, we employ multi-head attention, where $K$ independent attention mechanisms (heads) are applied in parallel, and their outputs are concatenated or averaged to form the final output.

The GCN/GAT layers are supplemented by a VNN to map the model's output to the socioeconomic target variable in a given region. The MSE objective function \ref{eq:1} from the VNN model is thus changed as $\frac{1}{N^2}\sum_{i}^{N}(Y_{i} - \hat{Y_{i}})^2$, where $Y_{i}, \hat{Y_{i}}$ are the ground truth and predicted target variable in question. This operation can be defined as follows: $\hat{Y_{i}} = f(h_{i}^{(2)})$. Here, $h_{i}^{(2)}$ corresponds to the vector embedding produced by the second GCN/GAT layer and $f$ is the mapping function (VNN). The model parameters $\theta_{GNN}$ and $\theta_{VNN}$ for GNN and VNN layers are learned in a single backpropagation pipeline. Notably, the learned embedding from the two-layer GCN/GAT $H^{(2)}$ is thus optimized for the specific purpose of socioeconomic modeling, ensuring that they capture essential network features conducive to accurately model a city's socioeconomic indicator.

\subsection{Experiments}
The broader aim of our experiments with mobility networks is to investigate their utility in socioeconomic modeling in a city. Graph-based modeling with mobility networks has demonstrated effectiveness in smaller urban networks, treating large urban regions (zipcodes) as nodes \cite{GNN_mobilityNet}. In this study, we extend our focus to larger urban networks, where smaller geographical units (census tracts) serve as nodes. This expansion in scale allows for the evaluation of the stability and consistency of mobility networks as predictors in more extensive urban areas characterized by complex mobility dynamics. As socioeconomic indicators, recent works have considered regional variables like crime statistics, personal income, bike flow, etc. \cite{NetRepr_GAT2}. We focus on median household income as the target variable in this work. Income is not only a direct measure of economic health in a neighborhood but also an indicator of development and investments in urban areas \cite{income_target}. Moreover, the income data is consistently available and easily accessible in urban areas, typically through census. 

With VNN-based embedding, a separate supervised learning model is considered to predict the target variable with the embedding as input. A VNN with (32, 64, 32) dimensional hidden layers is considered with Mean Squared Error (MSE) objective and Rectified Linear Unit (ReLU) activation, with the parameters being optimized with backpropagation. Proving the utility of mobility embeddings across all cities, we next experiment with training the GNN+VNN single pipeline architecture, further establishing the importance of network topology without any regional node features.

\section{Results}\label{sec: Results}

We present the R-squared (R2) scores corresponding to each configuration for median income modeling in Table~\ref{tab:ResultsTable}. The R2 scores being calculated as $R^2 = 1 - \frac{\sum_{i=1}^{n} (y_i - \hat{y}_i)^2}{\sum_{i=1}^{n} (y_i - \bar{y})^2}$, with results showing out of sample values where data was split in 70:30 ratio for model training/testing. Also shown are the R2 scores with 311 complaint features\cite{SocioeconomicModels3}, which were utilized in a MLP model for income modeling. For our proposed VNN based embedding model, we present the results with four different network embeddings as inputs - obtained from Spatial (location coordinates of nodes), SVD, Laplacian Eigenvectors (LE), and Random Walk methods. We notice the consistently superior results with the mobility network's embedding over the benchmark 311 features in 9 out of 12 cities in consideration, whereas 3 cities - Houston, San Antonio and Phoenix do not have 311 data available. In any case, we observe better modeling results from the proposed methods over local socioeconomic variables like job and population density.

\begin{table}[h]
\centering
\caption{\textbf{Out-of-sample R-2 values with the proposed methods for modeling median income across 12 U.S. cities}. The proposed models are able to outperform the models based on 311 features in cities. Four sets of network embeddings are considered as inputs to the models. Spatial and SVD embedding are found to be the most effective. The stability of results with embeddings (with margin of errors) is presented in Appendix:~\nameref{appendix:EmbedEx}. (\textbf{bold} represents best R2 score)}
\label{tab:ResultsTable}
\renewcommand{\arraystretch}{1.5} 
\resizebox{\textwidth}{!}{%
\begin{tabular}{|c!{\vrule width 1pt}c|*{5}{c|}}
\hline
\multicolumn{1}{|c|}{} & \multicolumn{1}{c|}{\textbf{Comparison Benchmark\cite{SocioeconomicModels3}}} & \multicolumn{4}{c|}{\textbf{VNN $|$ (GNN+VNN) $|$ (GAT+VNN)}} \\ \hline
\multicolumn{1}{|c|}{} & 311 features & Spatial & SVD & LE & Random Walk \\ \hline
\specialrule{.1em}{.05em}{.05em} 
NYC & 0.49 & 0.55 $|$ \textbf{0.58} $|$ 0.58 & 0.30 $|$ 0.46 $|$ 0.29 & 0.33 $|$ 0.20 $|$ 0.20 & 0.35 $|$ 0.45 $|$ 0.44  \\ \hline
LA & 0.16 & 0.13 $|$ \textbf{0.32} $|$ 0.28 & 0.11 $|$ 0.31 $|$ 0.31 & 0.1 $|$ 0.24 $|$0.24  & 0.1 $|$ 0.21 $|$ 0.20\\ \hline
Chicago & 0.59 & \textbf{0.7} $|$ 0.69 $|$ 0.68 & 0.3 $|$ 0.44 $|$ 0.44 & 0.5 $|$ 0.15 $|$ 0.14 & 0.56 $|$ 0.61 $|$ 0.60\\ \hline
Boston & 0.15 & 0.35 $|$ \textbf{0.50} $|$ 0.50 & 0.3 $|$ 0.28 $|$ 0.25 & 0.44 $|$ 0.14 $|$ 0.10 & 0.35 $|$ 0.21 $|$ 0.15\\ \hline
Philadelphia & 0.51 & 0.28 $|$ 0.33 $|$ 0.33 & 0.51 $|$ \textbf{0.55} $|$ 0.52 & 0.3 $|$ 0.31 $|$ 0.3 & 0.3 $|$ 0.45 $|$ 0.45 \\ \hline
Houston & NA & 0.23 $|$ 0.17 $|$ 0.15 & 0.36 $|$ \textbf{0.42} $|$ 0.40 & 0.19 $|$ 0.18 $|$ 0.18 & 0.25 $|$ 0.39 $|$ 0.39\\ \hline
Dallas & 0.37 & 0.33 $|$ 0.27 $|$ 0.26 & \textbf{0.58} $|$ \textbf{0.58} $|$ 0.56 & 0.38 $|$ 0.08 $|$ 0.05 & 0.36 $|$ 0.41 $|$ 0.40 \\ \hline
Austin & 0.28 & 0.43 $|$ 0.38 $|$ 0.38 & \textbf{0.59} $|$ 0.57 $|$ 0.53 & 0.33 $|$ 0.04 $|$ 0.07 & 0.36 $|$ 0.31 $|$ 0.34 \\ \hline
San Jose & 0.46 & 0.21 $|$ 0.4 $|$ 0.4 & \textbf{0.75} $|$ 0.42 $|$ 0.4 & 0.21 $|$ 0.21 $|$ 0.24 & 0.36 $|$ 0.03 $|$ 0.09 \\ \hline
San Diego & 0.36 & 0.16 $|$ 0.26 $|$ 0.28 & \textbf{0.43} $|$ 0.34 $|$ 0.32 & 0.27 $|$ 0.26 $|$ 0.25 & 0.36 $|$ 0.41 $|$ 0.33 \\ \hline
San Antonio & NA & 0.3 $|$ 0.48 $|$ 0.41 & \textbf{0.52} $|$ \textbf{0.52} $|$ 0.33 & 0.3 $|$ 0.04 $|$ 0.03 & 0.14 $|$ 0.34 $|$ 0.3 \\ \hline
Phoenix & NA & 0.14 $|$ 0.25 $|$ 0.25 & 0.15 $|$ \textbf{0.26} $|$ 0.25 & 0.21 $|$ 0.25 $|$ 0.24 & 0.15 $|$ \textbf{0.26} $|$ 0.25 \\ \hline
\end{tabular}
}
\end{table}

While spatial embeddings yield the strongest performance, we also observe that other types of network-derived embeddings exhibit substantial explanatory power in socioeconomic modeling. Importantly, our models do not incorporate any region-specific features; the embeddings are learned purely from the structure of the mobility network. This design choice is central to our objective: to evaluate whether mobility patterns alone—without any contextual or demographic information—are sufficient for downstream socioeconomic prediction tasks. By excluding explicit node features, we are able to directly assess the predictive value of network structure itself. The performance is improved with the (GNN+VNN) framework, which directly models the target socioeconomic variable. Improved results by stacking GNN layers make sense as Graph convolution and attention layers can capture network topology more intricately. However, incorporating attention with GAT does not improve the results, and in many cases achieves similar performance as with using just the convolution layers.
Evaluation with cross-validation indicates consistent and stable improvement of mobility networks over contextual node features as input.

\section{Discussion}

Our experiments across 12 cities show 1. the effectiveness of mobility networks in modeling median income and, consequently, 2. the ability of a GNN+VNN-based architecture to model a socioeconomic indicator in a single pipeline. 
While traditional network embedding methods fail to get meaningful representations for socioeconomic modeling, Vanilla Neural Network-based embeddings serve as good predictors in all cities. VNN-based embeddings, in a way, also capture network effects, as the training objective is designed to model the commute network weights in a city. 

\subsection{Interpretation of embedding configurations}
In many urban studies, initial configurations of spatial embeddings consistently show significantly better predictive performance compared to other approaches. Even when contrasted with local variables like job and population density, network embeddings demonstrate a superior ability to model income structures across cities. The embedding spaces inputted into our models capture distinct aspects of the network: SVD and Laplacian Eigenvectors (LE) emphasize structural and connectivity patterns, while random walk embeddings focus on connectivity and proximity. Although spatial embeddings may not directly represent network structure, tuning these embeddings within our models incorporates certain network characteristics after training. The superior performance of spatial embeddings in many cities suggest that inherent urban spatial residential patterns are critical in predicting median income. However, this does not discount the role of network interactions and structure, as other embeddings still outperform purely local variables. SVD embeddings likely excel because they capture core connectivity and community structure, which are important for modeling economic indicators. On the other hand, the relatively lower performance of random walk and Laplacian Eigenvector (LE) embeddings could be due to their focus on specific aspects of network connectivity that might not directly align with socioeconomic factors. Random walk embeddings prioritize proximity and frequently visited paths, which could lead to overemphasis on highly localized network areas, potentially missing broader spatial patterns relevant to income distribution. LE embeddings, while capturing global connectivity patterns, may introduce noise by highlighting structural nuances that aren’t as pertinent to income prediction. Further analysis comparing the significance of network structure versus residential location could be enhanced by modeling additional socioeconomic variables such as unemployment rates, housing availability, and more.

\subsection{Graph vs Non-Graph models}
Generally, the combination of GNN or GAT with VNN models outperforms standalone VNN models, suggesting that graph-based models capture spatial dependencies effectively. Furthermore, the GAT+VNN combination in the random walk feature consistently yields competitive performance, indicating that the attention mechanism in GATs effectively incorporates neighborhood information. However, there are cases where VNN-only models outperform the graph-based combinations, such as in Chicago and San Jose, suggesting that for certain cities, neighborhood relational data might be less informative for income modeling. This could be due to the particular urban layouts or spatial distributions in these cities. This, in fact, highlights that incorporating graph structures may or may not be useful depending on the spatial characteristics of the city. It is important to note that in certain cities, such as Los Angeles and Phoenix, even GNN-based models do not achieve strong predictive performance. This suggests that mobility patterns in these cities may differ substantially from those in other urban areas, and may not strongly correlate with socioeconomic indicators. Unique urban forms and spatial layouts likely contribute to this divergence. For example, the extensive urban sprawl of Los Angeles has been widely studied and contrasted with more compact city structures \cite{Chen2000ScienceOfSmartGrowth}. Similarly, the polycentric nature of Phoenix—characterized by multiple, loosely connected urban centers—has been shown to influence its distinct mobility behavior \cite{Leslie2006PolycentricPhoenix}. Such structural differences may weaken the predictive power of models relying solely on commute-based network information.

\paragraph{}
It's noteworthy that mobility networks as input exhibit superior performance compared to 311 data features and demographic variables such as population and job density, which represent comprehensive regional-level variables. When optimal network embedding is employed in conjunction with the 311 features as modeling input, notable improvements in results are observed in New York City and Chicago. However, incorporating population and job density alongside network embedding does not yield similar improvements (refer to Appendix:~\nameref{appendix:featureConcat}). It is also worth noting the significant results just based on network embedding, which shows that network-based representations are extensive enough to capture a lot of node information. We have focused on just the mobility network, however, it could be interesting to see how other network embedding compares to node features. GNN+VNN model can learn all parameters in a single gradient descent pipeline, hence the intermediate network representations (output of GNN layers) is learned with the objective of modeling socioeconomic target variable. This is in contrast to traditional representation learning, where network embedding vectors are typically learned with an objective based on network reconstruction.

\subsection{Policy implications and limitations}
The ability of mobility networks to effectively model socioeconomic indicators has significant policy implications. First, these models offer a scalable and data-efficient way to assess urban inequality by identifying structurally marginalized or disconnected regions based solely on commuting patterns. This can inform targeted interventions in under-served areas, even when fine-grained demographic or economic data is unavailable. Second, the framework enables planners to simulate and evaluate the potential socioeconomic impact of proposed infrastructure projects by modifying the network structure and observing the predictions of the downstream model. For instance, the addition of a transit link or road segment can be tested for its potential to improve modeled income predictions in peripheral neighborhoods. Finally, the end-to-end trainability of the model allows rapid retraining with updated mobility data, making it suitable for integration into real-time monitoring and response systems, especially useful in dynamic contexts such as post-disaster recovery, pandemic-related disruptions, or changes in commuter behavior. By learning directly from urban structure, the proposed GNN+VNN pipeline provides an interpretable and transferable tool for data-driven decision-making in urban policy.

While the results are promising, several limitations remain. First, the generalizability of our approach to cities outside the U.S. is yet to be validated, especially in regions with different urban forms, commuting patterns, or socioeconomic structures. And while census data is collected worldwide, the model's performance could be inherently sensitive to the quality and granularity of mobility data. In this regard, investigating the robustness of model with respect to inconsistent and inaccurate mobility data could be a follow-up research direction. Furthermore, although our deep learning framework offers strong predictive performance, its interpretability remains limited for non-technical stakeholders. Thus integrating interpretability mechanism in the model architecture offers another future direction. 

\section{Conclusion}

Urban networks have demonstrated strong modeling capabilities across a wide range of downstream tasks, such as traffic forecasting, land-use inference, and road-safety sensing \cite{NetworkInCities3,NetRepr_streetNetwork2, zhang2020multi}. In this work, we show that large-scale urban mobility networks—constructed from census-based commute flows—hold significant predictive power for socioeconomic modeling. Crucially, our models rely solely on the structure of the mobility network, without incorporating any node-level features, allowing us to isolate and evaluate the value of network topology alone.
We first demonstrate that node embeddings learned through a VNN trained to reconstruct the network can serve as powerful feature representations for urban areas. These embeddings outperform models trained on extensive handcrafted node features, achieving consistent improvements across all evaluated cities. Notably, our findings also echo broader trends in the literature, where deep learning–based embeddings have been shown to overcome limitations of traditional network embedding methods \cite{DeepLearningEmbedd, DeepLearningEmbedd2}.
Building on this, we propose a dedicated Graph Neural Network (GNN)-based model for socioeconomic prediction that jointly learns network structure and target modeling in a single, end-to-end pipeline. This unified approach eliminates the need for a two-step training process and directly optimizes all model parameters for the final prediction task. As a result, it not only improves efficiency but also delivers superior performance compared to both node feature–based models and separate embedding-plus-regression pipelines.

While the proposed methods show significant benefits of network-based modeling over contextual node features, more Graph-based deep learning methods could be potentially useful in homogeneous urban networks. Transformer architectures in graphs have been shown to achieve significant improvements in tasks concerning networks such as molecular structure networks \cite{GraphTransformers}. Additionally, many more urban networks could be evaluated for getting node representations \cite{NetworkInCities2, OtherUrbanNetworks, OtherUrbanNetworks2}, combining street networks, POIs, and building footprints. Moreover, other socioeconomic indicators like the housing profile of citizens and unemployment rate could also be modeled as a function of heterogeneous networks. Such models could help understand a complete socioeconomic picture of a neighborhood.

Urban scientists have long explored the prediction and analysis of socioeconomic status in cities using a wide range of social, economic, and spatial features \cite{SocioeconomicModels1, SocioeconomicModels2}. While these approaches have yielded valuable insights, they often overlook the structural interactions between regions—interactions that are naturally captured by urban mobility networks. Recent work has begun to explore the role of network-based representations in urban modeling, suggesting their potential for capturing latent spatial dynamics \cite{GNN_mobilityNet}. In our experiments in twelve major U.S. metro areas, we show that deep learning architectures have powerful modeling capabilities with mobility networks. Our work offers urban researchers a scalable and generalizable framework for incorporating inter-neighborhood interactions into their analyses. Beyond static modeling, the proposed methods can be adapted to dynamic or real-time mobility networks, offering tools for responsive urban planning and policy decision-making. Future researchers and practitioners should recognize the potential of mobility-based network signals as standalone predictors, and consider integrating such structural representations into broader frameworks for socioeconomic forecasting, urban resilience planning, and real-time intervention strategies.

\paragraph{Acknowledgments}
This research was supported by the MUNI Award in Science and Humanities (MASH Belarus) of the Grant Agency of Masaryk University under the Digital City project~(MUNI/J/0008/2021).

\paragraph{Data Availability}
The data used in this study can be found in a Zenodo repository: \url{https://doi.org/10.5281/zenodo.11494208}.
Specifically, the data contains origin-destination daily commute flow information among the census tracts in all the 12 U.S. cities considered in this work.

\paragraph{Author Contributions}Conceptualization, D.K. and S.S.; methodology, D.K. and S.S.; formal analysis, D.K.; investigation, D.K. and A.B.; data curation, D.K.; writing---original draft preparation, D.K.; writing---review and editing, A.B. and S.S.; visualization, D.K.; supervision, S.S.; project administration, S.S. and A.B.; funding acquisition, S.S. All authors have read and agreed to the published version of the manuscript.

\paragraph{Conflicts of interest}
The authors declare no conflicts of interest.

\vspace{6pt}




\appendix

\section{}

\subsection{Data}\label{appendix:Data}

The mobility information for Chicago and Boston was retrieved from the LEHD \cite{LEHDData}. The 311 complaint data set for the two cities is available from the official city data portals \cite{Boston311, Chicago311}. The 311 data set is comprehensive data of the non-emergency complaints from citizens directed at various city agencies. Fig.~\ref{fig:311_categories} shows the complaints distribution by top categories in each city.

\renewcommand{\thefigure}{A1}
\begin{figure}[h]
    \centering
    \subfigure[NYC]{
        \includegraphics[width=0.4\textwidth]{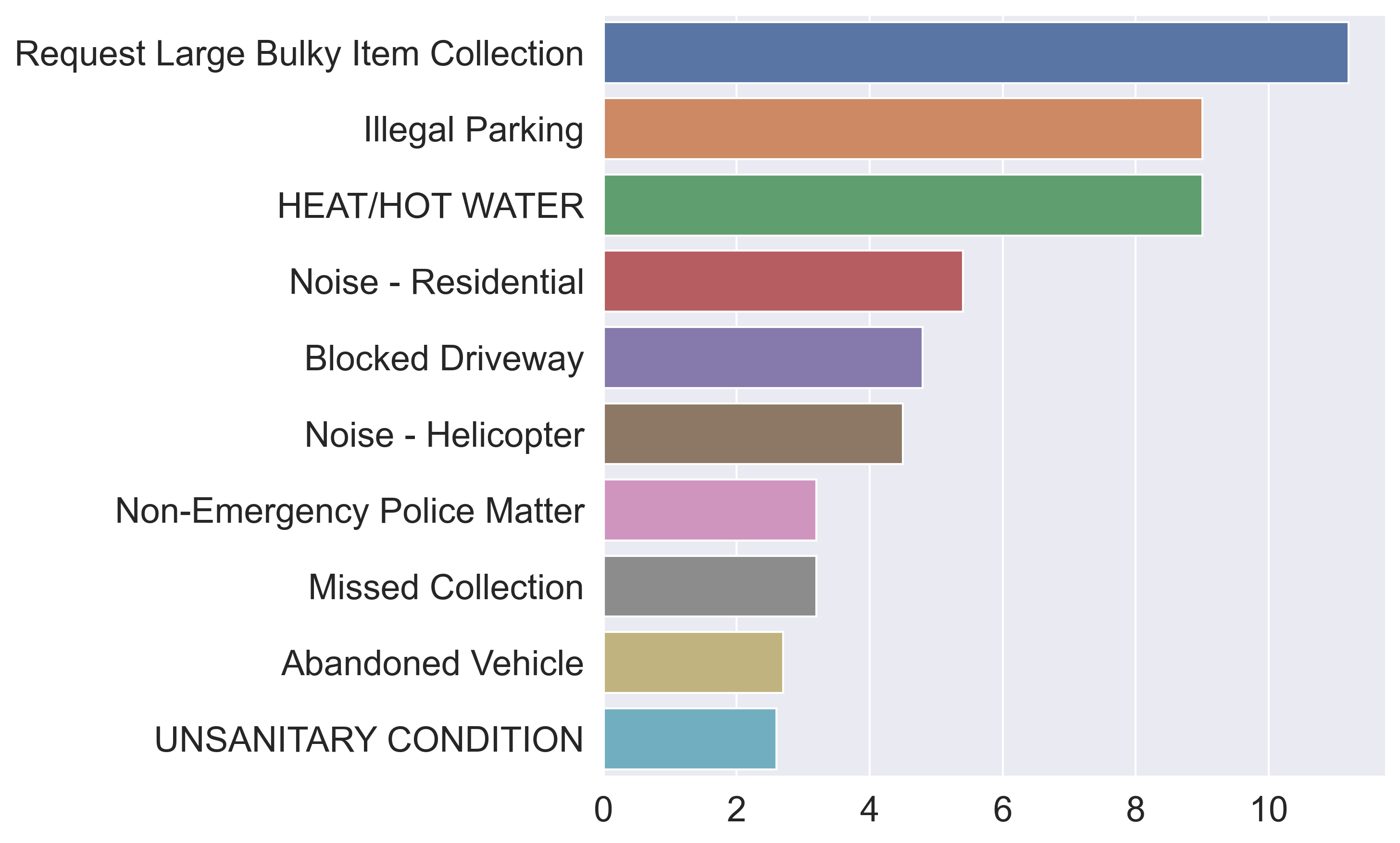}
        }
    \subfigure[Chicago]{
        \includegraphics[width=0.4\textwidth]{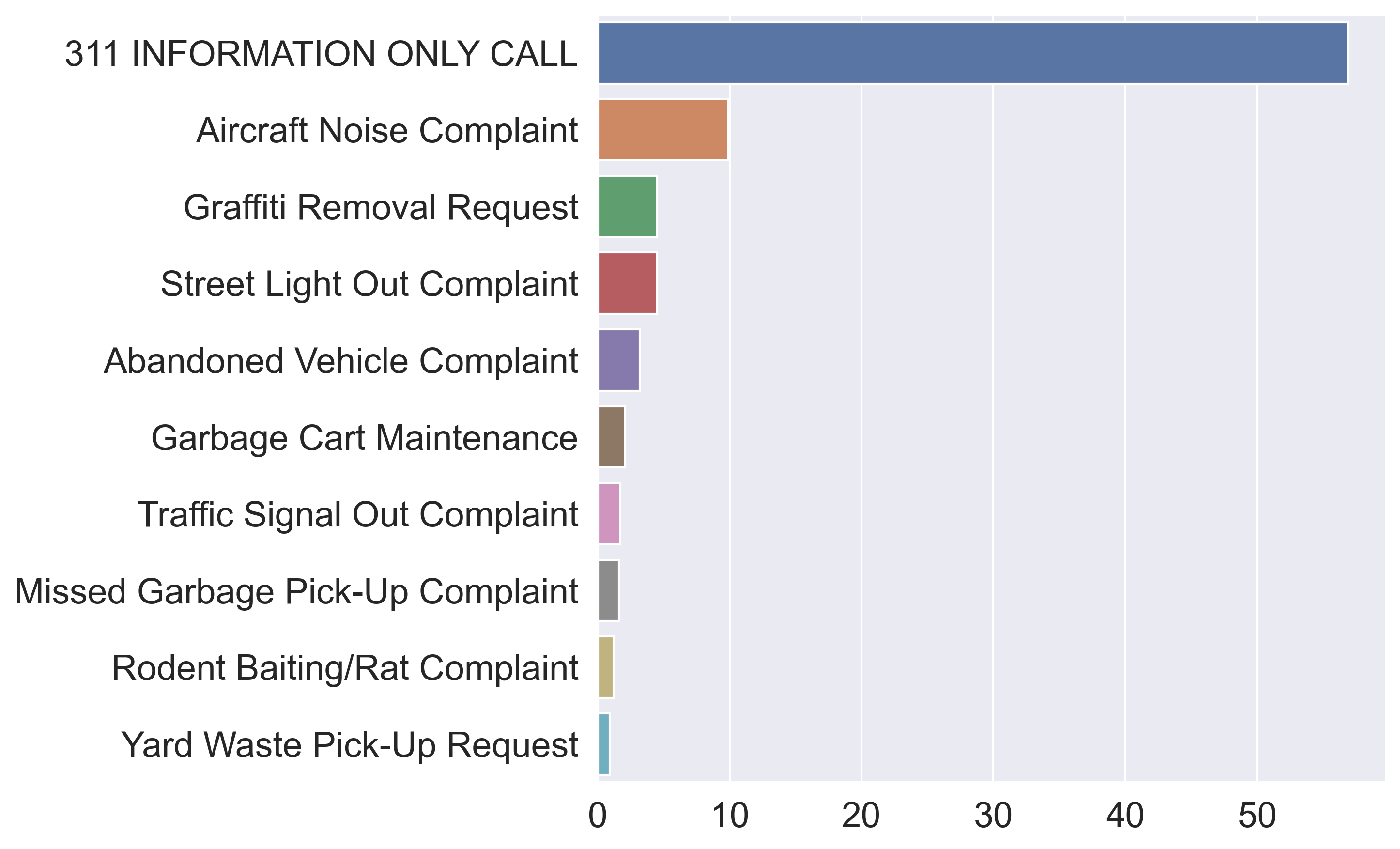}
        }
    \subfigure[Boston]{
        \includegraphics[width=0.4\textwidth]{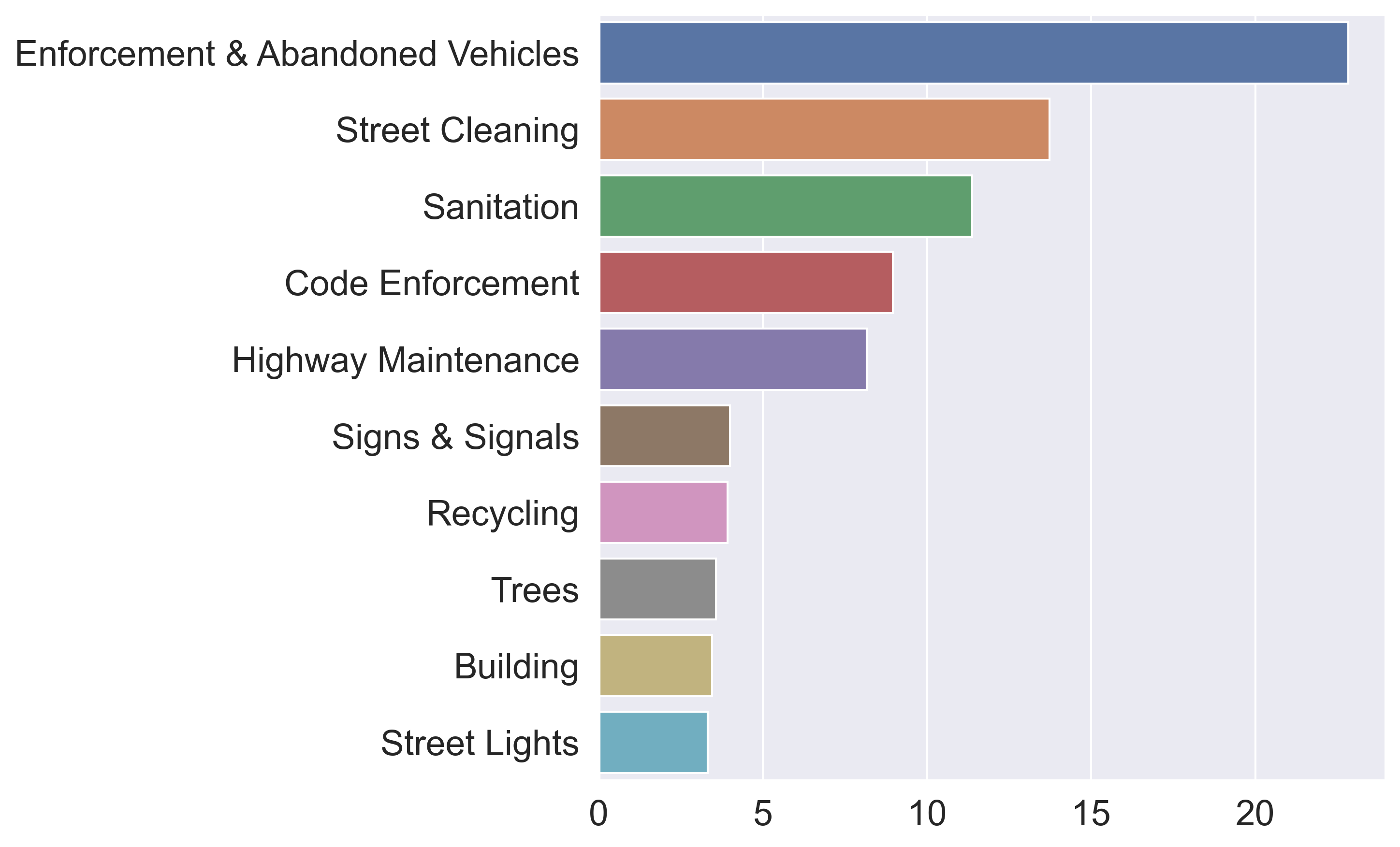}
        }
    \caption{Top 10 complaint categories in the 311 data, represented as \% of total complaints, across each city}
    \label{fig:311_categories}
\end{figure}

In modeling, the 311 data is aggregated at the census tract level and normalized across all complaint categories to represent the proportion footprint of each category in a location. This approach was proposed in \cite{SocioeconomicModels3} to model income, and we use the same pre-processing for 311 data to use as inputs in our models. 

We also consider two more demographic variables - population density and job density, as modeling inputs for comparison purposes. These features are available from the U.S. census \cite{ACSData} and aggregated to the census tract level for our study. Fig.~\ref{fig:population_density} shows population density across the three cities.

\renewcommand{\thefigure}{A2}
\begin{figure}[h]
    \centering
    \subfigure[NYC]{
        \includegraphics[width=0.42\textwidth]{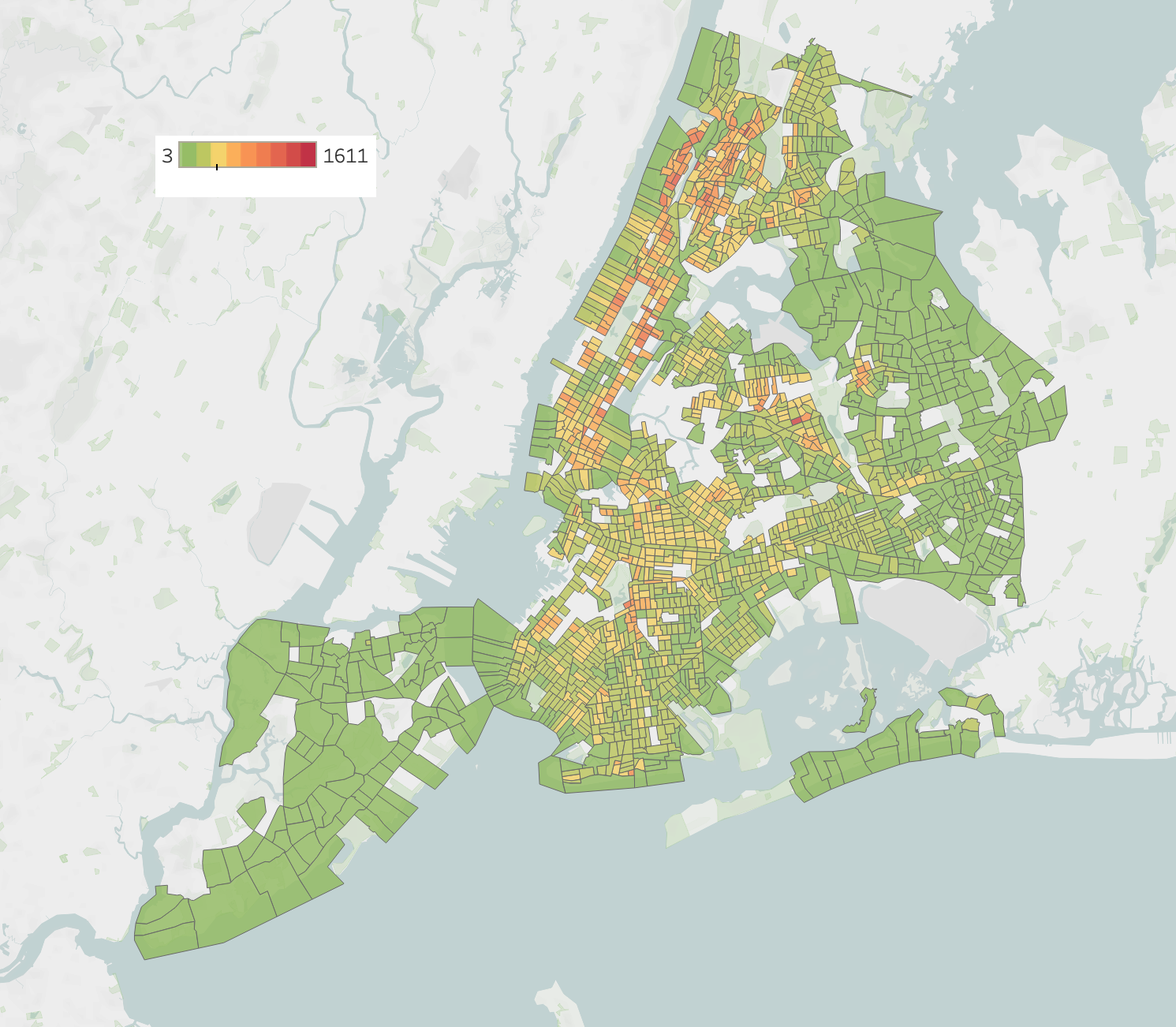}
        }
    \subfigure[Chicago]{
        \includegraphics[width=0.39\textwidth]{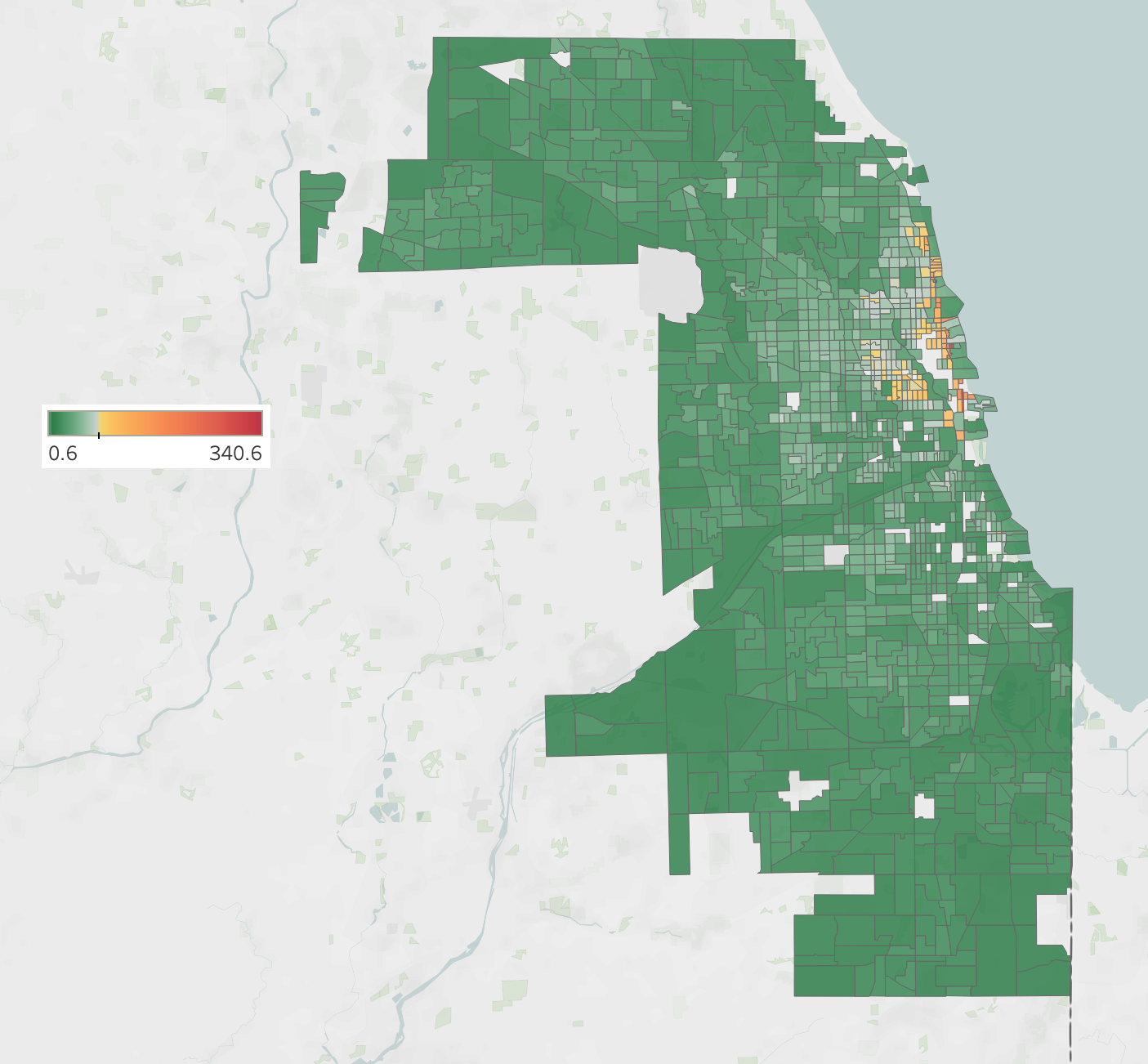}
        }
    \subfigure[Boston]{
        \includegraphics[width=0.37\textwidth]{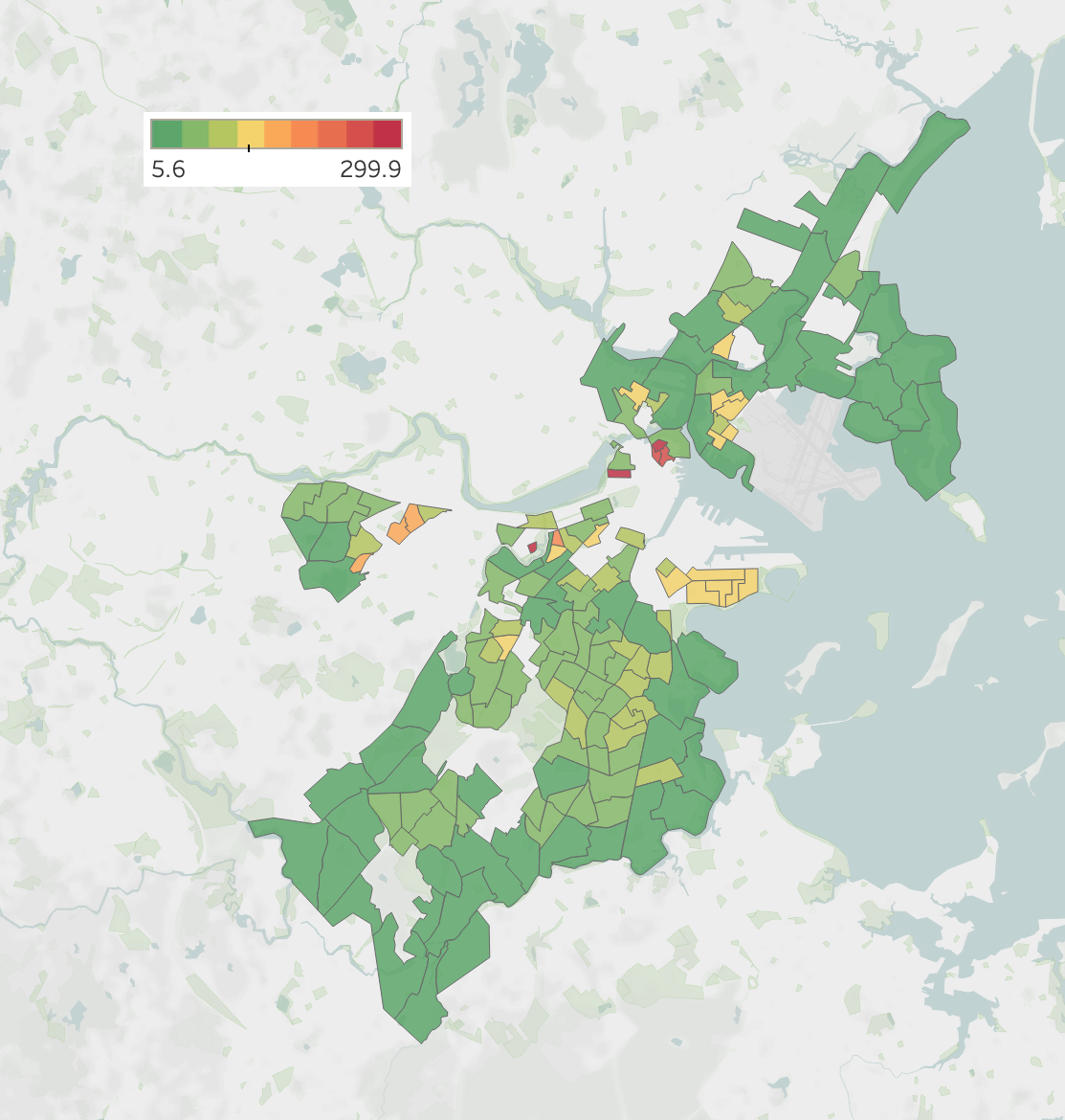}
        }
    \caption{Population density (per 100 sq.meters) across census tracts in cities}
    \label{fig:population_density}
\end{figure}

\subsection{Network embedding clustering}\label{appendix:EmbeddingViz}

With the idea of identifying spatial patterns (if any) related to socioeconomics, we experimented with clustering mobility embeddings derived from Random Walk and Laplacian Eigenvectors (LE) methods. We employ Pagerank method \cite{Pagerank} as a Random Walk embedding, owing to its success in urban networks in the literature \cite{UrbanPagerank1, UrbanPagerank2}. Results from K-means clustering method are shown in Fig.~\ref{fig:pagerankViz} and Fig.~\ref{fig:LapViz} for Pagerank and LE embeddings respectively.

\renewcommand{\thefigure}{A3}
\begin{figure}[h]
    \centering
    \subfigure[NYC]{
        \includegraphics[width=0.47\textwidth]{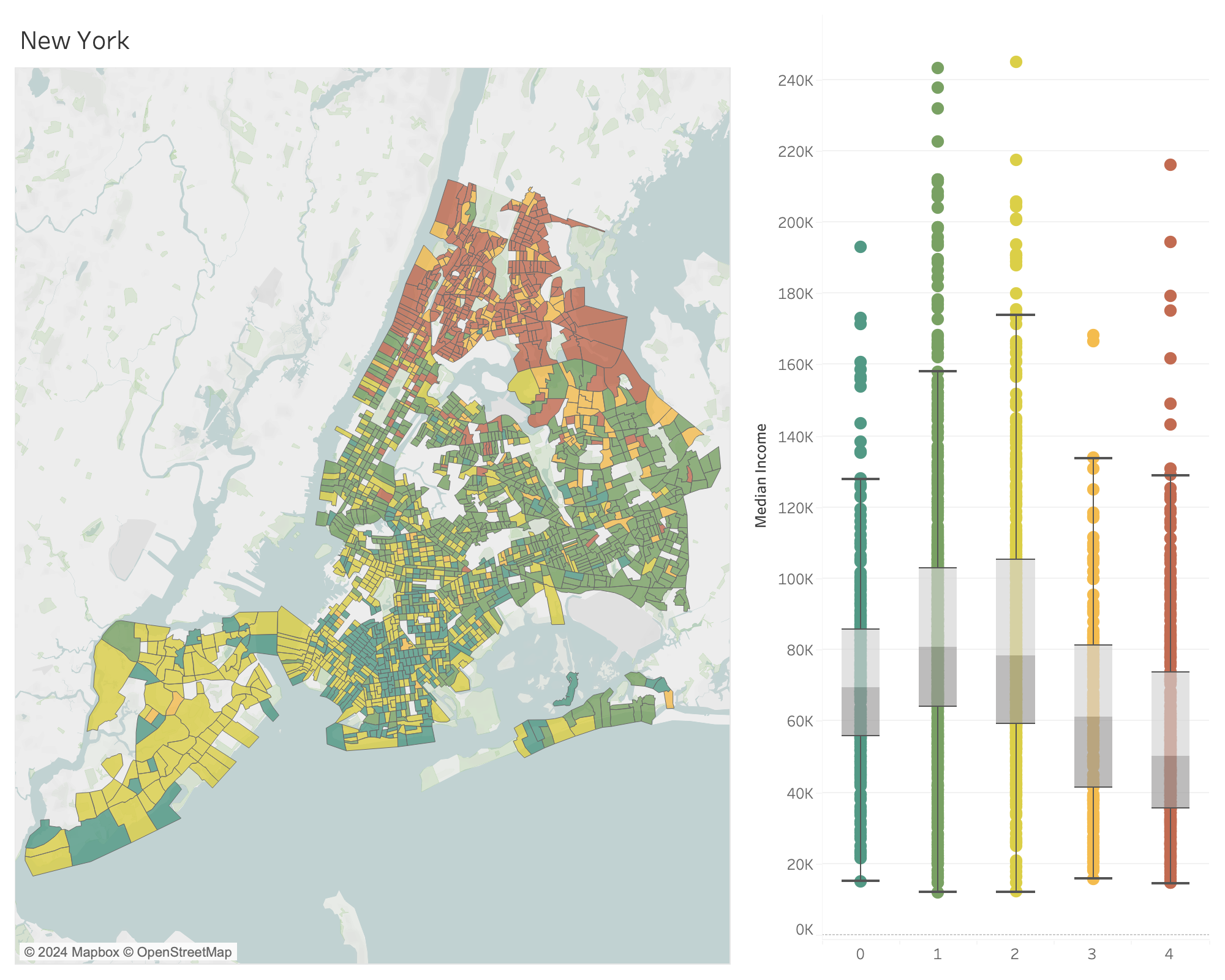}
        }
    \subfigure[Chicago]{
        \includegraphics[width=0.47\textwidth]{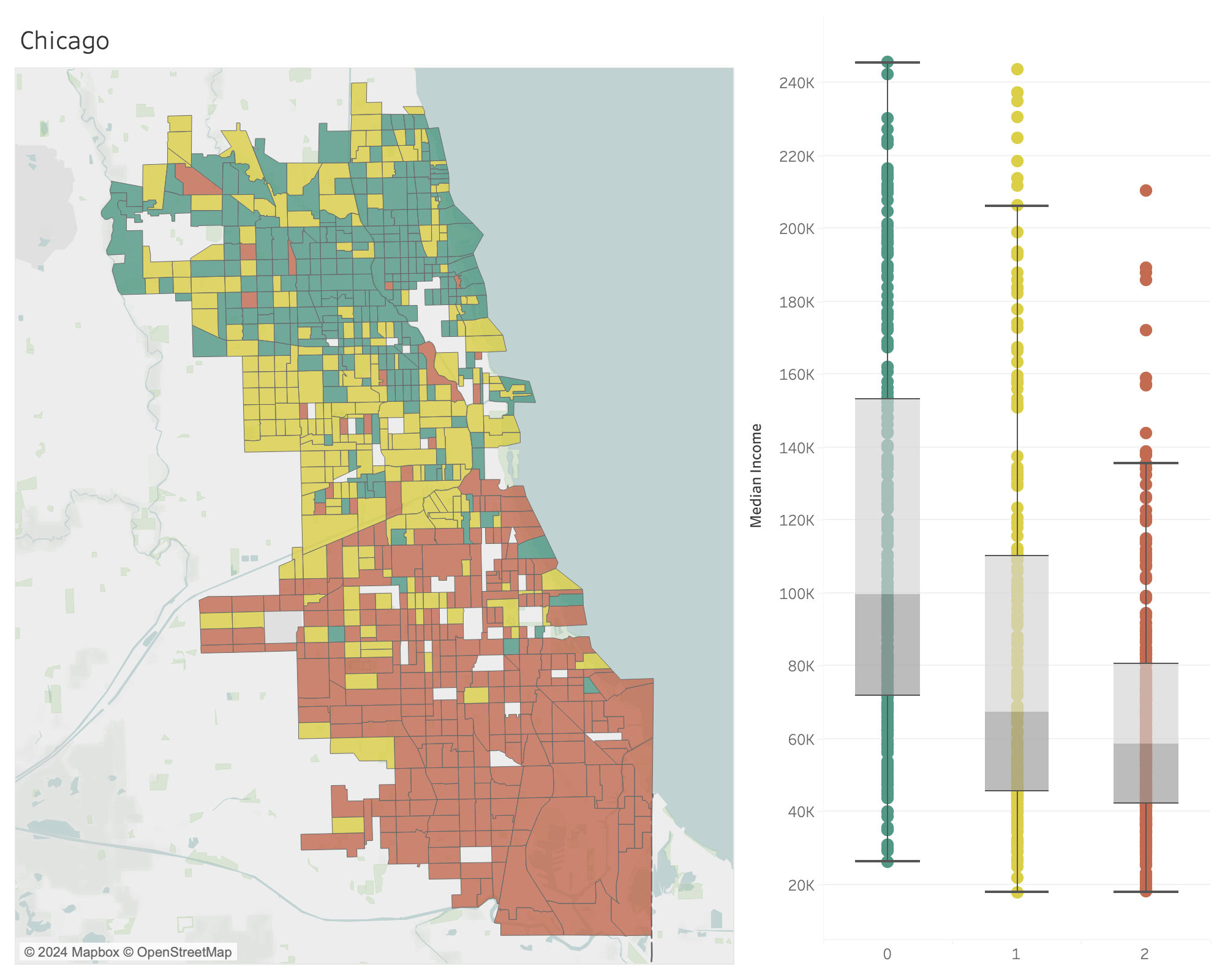}
        }
    \subfigure[Boston]{
        \includegraphics[width=0.45\textwidth]{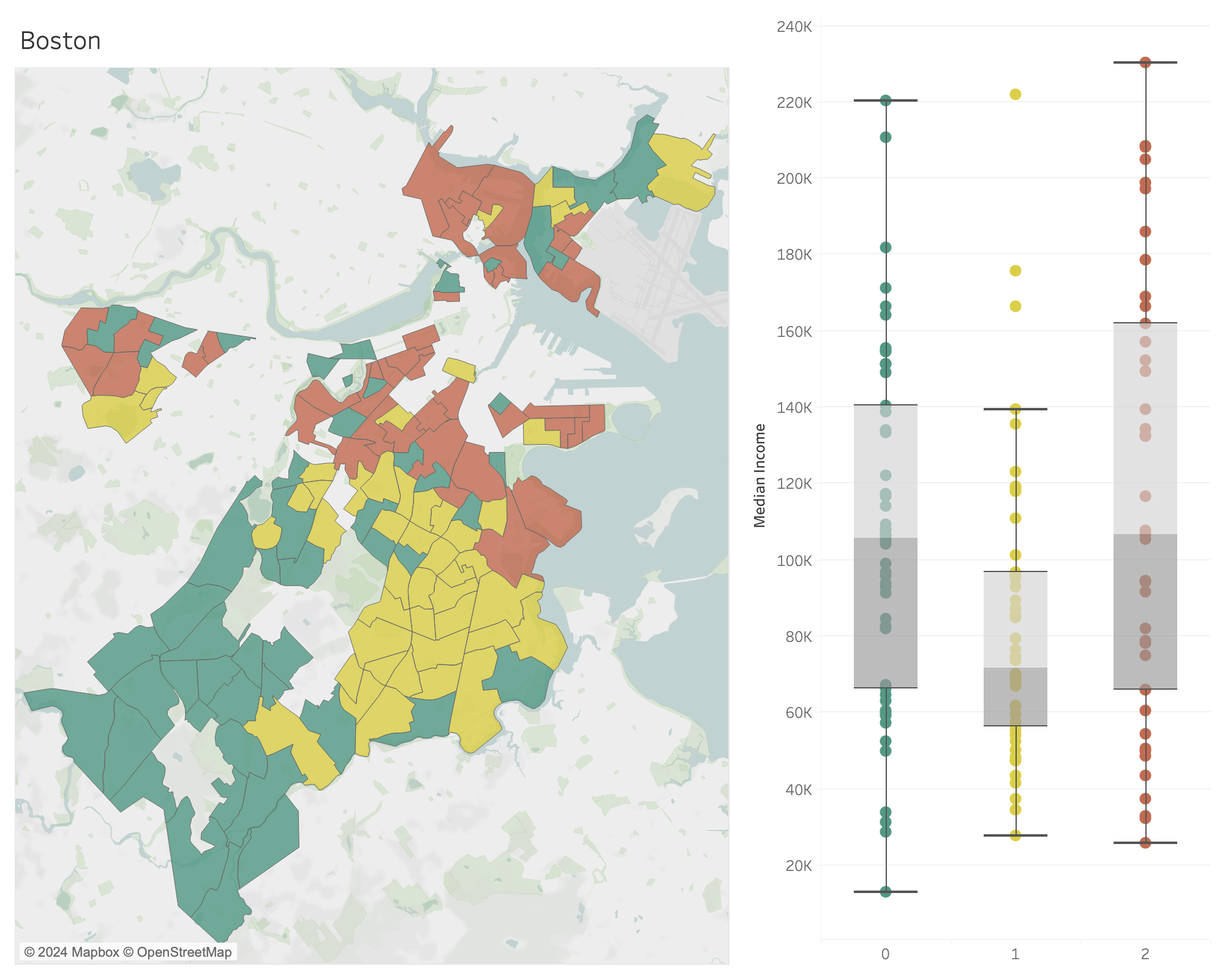}
        }
    \caption{Clustering of random walk (Pagerank) embedding of mobility networks - like SVD, we get distinction of neighborhoods based on income profiles}
    \label{fig:pagerankViz}
\end{figure}

\renewcommand{\thefigure}{A4}
\begin{figure}[h]
    \centering
    \subfigure{
        \includegraphics[width=0.47\textwidth]{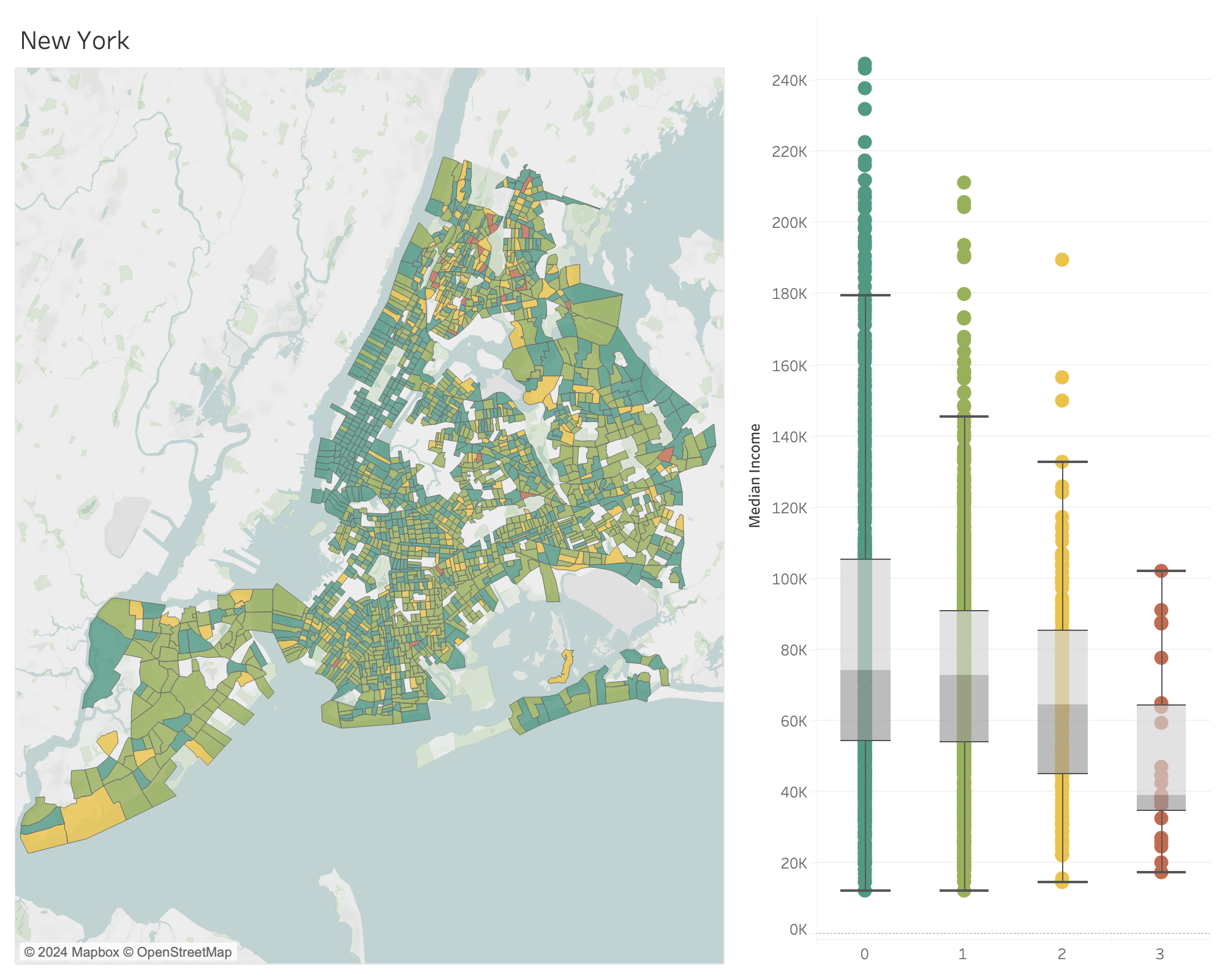}
        }
    \subfigure{
        \includegraphics[width=0.47\textwidth]{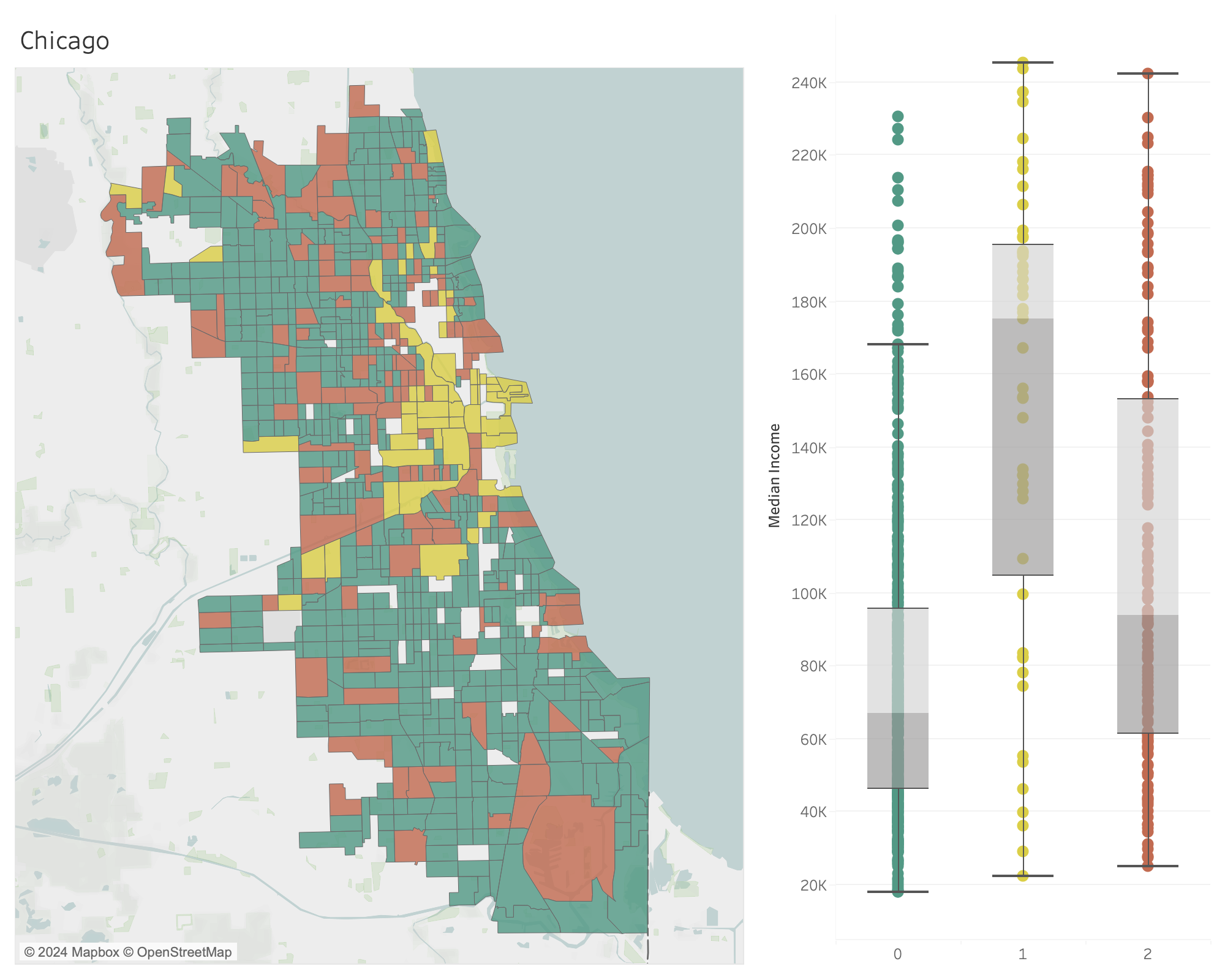}
        }
    \subfigure{
        \includegraphics[width=0.45\textwidth]{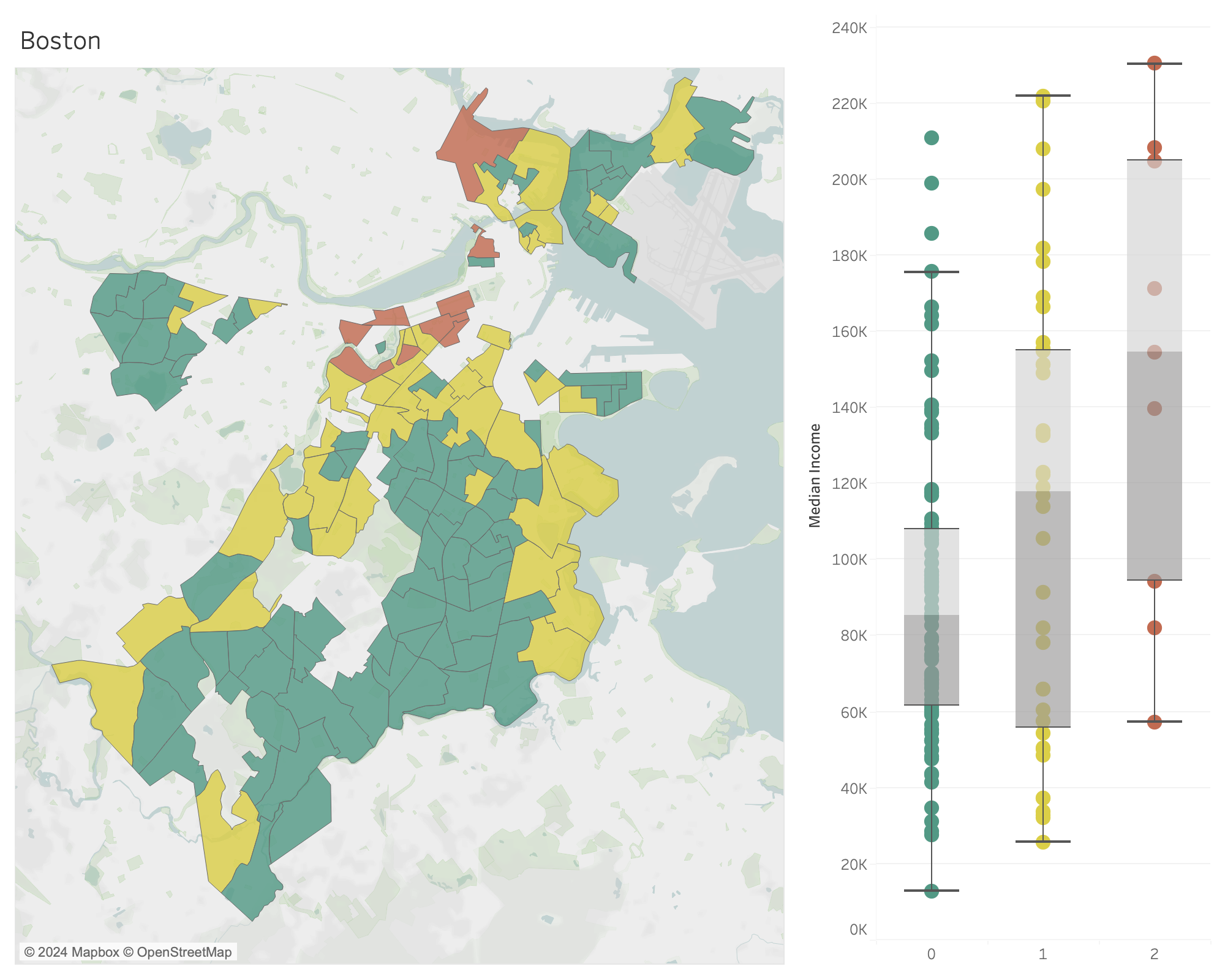}
        }
    \caption{Clustering of Laplacian embedding of mobility networks - while results are not so spatially cohesive as SVD and Pagerank methods, we can still distinguish high-income neighborhoods from the rest}
    \label{fig:LapViz}
\end{figure}

Like SVD, the Pagerank embedding for census tracts can distinguish the regions based on their income profile. For all the three cities in consideration, we see distinct neighborhoods identified by their differences in median income. 
However, clustering results from LE embedding is not so discernable. While we still see spatial patterns in clusters, the differences in income profile is not quite significant among most clusters in all cities. However, LE embedding can still distinguish some high-income neighborhoods from others. For instance, it can distinguish the lower Manhattan area of NYC and central Chicago as separate clusters. 

\subsection{Experiments with embedding: Dimensionality}\label{appendix:EmbedEx}

We experimented with different node embeddings dimensions to evaluate modeling median income at the node (census tract) level. Table \ref{tab:VNNembedding_dims} shows the R-2 scores in the three cities corresponding to each VNN-based embedding configuration. Note that the embeddings are used as inputs to a separate VNN model to model the target socioeconomic variable. 


\renewcommand{\thetable}{S1}
\begin{table}[H]
    \centering
    \renewcommand{\arraystretch}{1.5}
    \caption{\textbf{Embedding dimensions and stability} -- $R^2$ scores from modeling median income, with $d$-dimensional VNN-based embedding as modeling input.}
    \label{tab:VNNembedding_dims}
    \resizebox{\textwidth}{!}{
    \begin{tabular}{ccccccccccc}
        \toprule
        \textbf{VNN Embedding} & \textbf{NYC} & \textbf{Boston} & \textbf{Chicago} & \textbf{San Jose} & \textbf{San Diego} & \textbf{Austin} & \textbf{Dallas} & \textbf{LA} & \textbf{San Antonio} & \textbf{Phoenix} \\
        \midrule
        $d=2$  & $0.47 \pm 0.04$ & $0.03 \pm 0.02$ & $0.3 \pm 0.05$ & $0.47 \pm 0.06$ & $0.36 \pm 0.03$ & $0.1 \pm 0.04$ & $0.09 \pm 0.01$ & $0.13 \pm 0.02$ & $0.23 \pm 0.04$ & $0.16 \pm 0.05$ \\
        $d=5$  & $0.40 \pm 0.01$ & $0.32 \pm 0.01$ & $0.68 \pm 0.015$ & $0.63 \pm 0.03$ & $0.43 \pm 0.01$ & $0.59 \pm 0.03$ & $0.58 \pm 0.025$ & $0.28 \pm 0.01$ & $0.49 \pm 0.03$ & $0.44 \pm 0.01$ \\
        $d=10$ & $0.53 \pm 0.02$ & $0.18 \pm 0.01$ & $0.68 \pm 0.01$ & $0.75 \pm 0.02$ & $0.41 \pm 0.02$ & $0.54 \pm 0.03$ & $0.54 \pm 0.02$ & $0.28 \pm 0.015$ & $0.49 \pm 0.025$ & $0.45 \pm 0.01$ \\
        $d=15$ & $0.53 \pm 0.01$ & $0.26 \pm 0.005$ & $0.63 \pm 0.02$ & $0.70 \pm 0.03$ & $0.41 \pm 0.01$ & $0.54 \pm 0.03$ & $0.58 \pm 0.005$ & $0.26 \pm 0.01$ & $0.52 \pm 0.01$ & $0.42 \pm 0.005$ \\
        \bottomrule
    \end{tabular}
    }
\end{table}

Notably, the 5 dimensions are optimal for many of the cities, whereas, for NYC, San Jose and Phoenix, the optimal embedding dimension is 10. In San Antonio, we notice a slight improvement with increasing the dimensionality to 15. However, any larger dimensions don't contribute to more improvement. Moreover, the results prove be stable with the embeddings in all cities, with relatively low margin of errors.

\subsection{Concatenation and modeling with neighborhood-level features} \label{appendix:featureConcat}

To evaluate the combined modeling capability of both node embeddings and neighborhood-level features, we experimented with using them together as inputs to a supervised VNN model. Specifically, we concatenate the optimal VNN-based embedding with the 311 complaint feature set, population density, and job density of the neighborhoods. Table ~\ref{tab:featureConcat} shows results for 3 major cities.

\renewcommand{\thetable}{S2}
\begin{table}[H]
    \centering
    \begin{tabular}{cccc}
        \toprule
        \multicolumn{1}{c}{\textbf{Model inputs}} & \multicolumn{3}{c}{\textbf{Cities}} \\
        \cmidrule(lr){2-4}
        & \textbf{NYC} & \textbf{Boston} & \textbf{Chicago} \\
        \midrule
        \textbf{Population density} & 0.02 & 0.01 & 0.16 \\
        \textbf{Job density} & 0.07 & 0.08 & 0.12 \\
        \textbf{Embedding+Population density} & 0.53 & 0.33 & 0.68 \\
        \textbf{Embedding+Job density} & 0.53 & 0.33 & 0.68 \\
        \textbf{Embedding+311 data} & 0.64 & 0.33 & 0.75 \\
        \bottomrule
    \end{tabular}
    \caption{R-2 scores from across various input features to a supervised model to predict median income}
    \label{tab:featureConcat}
\end{table}



\bibliographystyle{unsrt} 
\bibliography{references}

\end{document}